# Hierarchical graph sampling based minibatch learning with chain preservation and variance reduction

Qia Hu [1,2†], Bo Jiao [1*†]

hqqqq@stu.xmu.edu.cn jiaoboleetc@outlook.com

[1] *Key Laboratory of Intelligent Manufacturing and Industrial Internet Technology, Fujian Province University, Xiamen University Tan Kah Kee College, Zhangzhou, 363123, Fujian, China.*
[2] *School of Informatics, Xiamen University, Xiamen, 361102, Fujian, China.*

**Abstract:** Graph sampling-based Graph Convolutional Networks (GCNs) decouple sampling from forward and backward propagation during minibatch training, enhancing scalability with respect to layer depth and graph size. We propose $HIS_{GCNs}$, a hierarchical importance sampling-based learning method. By constructing minibatches using sampled subgraphs, $HIS_{GCNs}$ focuses on the importance of both the core and periphery in a scale-free training graph. Specifically, it preserves the centrum of the core in most minibatches, which maintains connectivity between periphery nodes, and samples periphery edges without core node interference, which allows longer chains composed entirely of low-degree nodes remain within the same minibatch. $HIS_{GCNs}$ can maximize the discrete Ricci curvature (i.e., Ollivier-Ricci curvatures) of the edges in a subgraph, enabling preservation of important chains for information propagation. This approach can achieve a low node embedding variance and a high convergence speed. Diverse experiments on Graph Neural Networks (GNNs) with node classification tasks confirmed the superior performance of $HIS_{GCNs}$ in terms of both accuracy and training time.

## 1 Introduction

Graph Convolutional Networks (GCNs) perform efficient feature extraction and aggregation on attributed graphs [1,2]. However, with increasing scale of the graphs [3], GCNs face challenges owing to computation and storage complexity. One prominent issue is known as "neighbor explosion" [4,5]. In a GCN, feature to be gathered for a node originates from its neighbors in the previous layer, and each of these neighbors recursively gathers feature from its previous layer. For a scale-free graph with low diameter, the neighborhood of a single node expands rapidly, covering a substantial portion of the graph, even with a small batch size [5,6]. Graph sampling based GCNs resolve the "neighbor explosion" by building every GCN on a subgraph with small size [3,6].

Graph sampling based GCNs first extract $n$ subgraphs to construct minibatches, and then build a full GCN for each minibatch on a subgraph, yielding new challenges in training, such as bias to nodes frequently sampled and loss of long chains that help propagate long-distance features. Therefore, loss normalization is an important technology [6], while preserving critical chains in the same subgraph is indispensable, especially for long chains composed entirely of low-degree nodes, because chains with length two or greater represent the correlation between edges, and two connected nodes that have few neighbors are likely to be influential to each other [6]. Thus, we investigate a hierarchical importance graph sampling aiming to reduce node aggregation variance while preserving critical long chains in the same subgraph.

Many graphs, such as social, collaborative and biological networks, are scale-free. The topology of these graphs has a core-periphery structure caused by preferential attachment [7], which pref-erentially attaches each newly added node to high-degree nodes. Specifically, the structure is composed of a dense core with few high-degree nodes and a sparse periphery with massive low-degree nodes [8-12]. Based on preferential attachment, as the graph scale increases, the internal connections of the small and dense core are relatively stable, whereas the number of periphery nodes and their edges connected to the core grow rapidly. Thus, preserving the centrum structure of the small core is helpful for maintaining the connection of subgraphs with small size [12]. Periphery is sparse; however, its node number is large, resulting in a large number of long chains being contained within it. These chains, composed entirely of low-degree nodes, are critical for feature propagation [6]. Thus, the complete chains should be preserved as much as possible in the same minibatch, even when it is small. The training of GCNs relies on various minibatches, all of which help alleviate the information loss of these critical chains. The core-periphery structure enables each minibatch to simultaneously achieve two goals: preserving the core centrum to maintain connectivity, and preserving the peripheral long chains without core node interference. Therefore, we design a graph sampling based learning method $HIS_{GCNs}$ using the hierarchical core-periphery structure.

* Corresponding author.

† These authors contributed equally to this work.

Our contributions can be summarized as follows: 1. We propose a hierarchical importance graph sampling-based GCN minibatch learning method $HIS_{GCNs}$, objective to preserve critical long chains in the same minibatch and reduce node aggregation variance, 2. We performed theoretical and experimental analyses of chain preservation and variance reduction in minibatch training, and 3. We validated the effectiveness of $HIS_{GCNs}$ through extensive experiments on semi-supervised node classification tasks across various datasets, demonstrating their superior performance in terms of both accuracy and training time. Open-source code: *https://github.com/HuQiaCHN/HIS-GCN*.

The structure of this paper is organized as follows: Section 2 introduces related work, Section 3 provides design principles, Section 4 describes the implementation detail of $HIS_{GCNs}$, and Section 5 validates the effectiveness of chain preservation and variance reduction as well as demonstrates the superior performance of $HIS_{GCNs}$ in terms of both accuracy and training time.

## 2 Related work

### 2.1 Graph learning

Graph learning (i.e., machine learning on graphs) is highly effective for tasks, such as classification, link prediction, and matching. Related technologies include graph signal processing, matrix factorization, random walk, active learning, and deep learning, among others [13-15]. Deep learning, particularly Graph Neural Networks (GNNs), performs end-to-end learning through message passing and recursive aggregation of neighbor features. GNNs have diverse architectures, such as GCNs [1], GraphSAGE [4], and attention-based GAT [16]. Recently, sampling methods in deep learning have been widely used to improve aggregation capacity and scalability. Active learning aims to employ a few, high-quality samples as possible to achieve the best possible performance of the model [15]. Deep active learning has become an important research field [17]. In [18], the authors integrated multi-scale GraphSAGE with superpixel-based subgraph sampling for hyperspectral image classification. In [19], the authors designed a time-aware dynamic sampling strategy with attention gating to improve neighbor selection in autonomous driving graphs. In [20], the authors introduced a heterogeneity-aware sampling method that ranks neighbors by cosine similarity and greedily expands subgraphs for better coverage. These works involve deep learning with dynamic, heterogeneous, or task-specific graph structures. This paper studies sampling-based deep learning on static and homogeneous graphs that have received widespread attention.

## 2.2 GCN framework

GCNs extend convolution operations to graph-structured data, and propagate information between nodes in a graph [1-6]. For an un-directed, attributed graph $G = (\mathcal{V}, \mathcal{E})$ with nodes $v \in \mathcal{V}$ and edges $(u, v) \in \mathcal{E}$, where the number of nodes is $|\mathcal{V}| = N$ and the number of edges is $|\mathcal{E}|$. Each node in the graph is represented by a feature vector of length $f$. Subsequently, an adjacency matrix $A \in \mathbb{R}^{N\times N}$ and a feature matrix $X \in \mathbb{R}^{N\times f}$ can be defined, where $A_{u,v} = 1$ if there is an edge between nodes $u$ and $v$, and $0$ otherwise; the $i^{th}$ row of $X$ represents the feature vector of node $v_i$. Let $\widehat{A} = \widetilde{D}^{-1/2} \widetilde{A} \widetilde{D}^{-1/2}$, where $\widetilde{A} = A + I_N$, $I_N$ is an identity matrix and $\widetilde{D}$ is the diagonal degree matrix of $\widetilde{A}$ with $\widetilde{D}_{ii} = \sum_j \widetilde{A}_{ij}$. Let $W^{(l)} \in \mathbb{R}^{f^{(l)}\times f^{(l+1)}}$ be the trainable weight matrix at layer $l$, $\sigma$ be the activation function, and $H^{(l)} \in \mathbb{R}^{N\times f^{(l)}}$ be the node embedding matrix at layer $l$ with $H^{(0)} = X$. The architecture of GCNs can be summarized as follows:

$$H^{(l+1)} = \sigma\left(\widehat{A}\, H^{(l)} W^{(l)}\right) \tag{1}$$

Let $\widehat{A}_{v,u} \in \mathbb{R}$ be an element of $\widehat{A}$, and $H^{(l)}(v) \in \mathbb{R}^{f^{(l)}}$ be the feature vector of node $v$ at layer $l$. GCNs can be interpreted by a message-passing framework, where each node passes information to its neighboring nodes (the activation function is omitted):

$$H^{(l+1)}(v) = \sum_{u\in\mathcal{V}} \widehat{A}_{v,u}\, H^{(l)}(u) W^{(l)}, v \in \mathcal{V} \tag{2}$$

GCNs recursively aggregate information over multiple layers to generate representative feature vectors for each node and perform well in tasks, such as node classification. However, the computational complexity of Eq. (1) is $O(|\mathcal{E}| f^{(l)} f^{(l+1)})$, thereby hindering GCN training for large or dense graphs. In addition, both forward and backward propagation recursively expand node neighbors, which requires all nodes and edges to be loaded into memory, resulting in a huge memory requirement. A potential solution is to sample the training graph to reduce its scale; thus, reducing computational and memory complexity [3,6,21-32].

## 2.3 Sampling-based graph learning

### 2.3.1 Node-wise sampling methods

Node-wise sampling methods extract a subset of nodes and their neighbors in each convolution layer to reduce computational complexity. A representative method is GraphSAGE [4], generating node embeddings by randomly sampling a subset of *k*-hop neighbors. This addresses the issue of exponentially growing neighbors. In [22], the authors integrated efficient random walks and graph convolutions to generate embeddings of nodes that incorporate both graph structure and node feature information. In [23], the authors extended GraphSAGE by limiting only two neighbors of each node to be sampled and employed historical activations to reduce the variance. These methods can reduce memory demand by sampling neighbors for each forward propagation.

### 2.3.2 Layer-wise sampling methods

Layer-wise sampling methods demonstrated improvements over node-wise sampling methods. Instead of sampling neighbors for each node, these methods sample a fixed number of nodes in each convolution layer to further reduce computational complexity. FastGCN [24] is a representative method that executes independent batch sampling of nodes in each layer, thereby addressing the exponential growth of neighbors with increasing layers. Inter-layer dependence models have been proposed to address the issue of broken connectivity caused by independent sampling across layers. In [25], the authors designed a top-down sequential model in which sampling in the lower layers depends on the results of the higher layers. In [26], the authors designed a layer-dependent importance sampling model that ensures connectivity between the sampled nodes. Owing to the inter-layer dependence in sampling, these models incur additional time costs.

#### 2.3.3 Graph sampling based methods

Graph sampling based methods create many minibatches, each built on a subgraph that is composed of certain nodes and edges extracted from the same large-scale graph, and then train GCN on each of these minibatches and normalize correlation loss between different subgraphs. Cluster-GCN [27] partitions the graph into multiple clusters, and randomly combines a fixed number of clusters to a subgraph. RWT [3] samples subgraphs from the graph to constitute a mini-batch, and a full GNN is updated based on the mini-batch gradient. GraphSAINT [6] employs lightweight sampling algorithms, including importance-based random node/edge, random walks, and multi-dimensional random walks. In addition, it utilizes normalization techniques to balance the frequency of sampled nodes and edges across different subgraphs. SHADOW-GNN [28] decouples the depth and scope of GNNs to improve expressivity and scalability without modifying layer architecture, where the scope is related to a shallow subgraph. BNS-GCN [29] considers the cost of data transfer between GPU and CPU for boundary nodes of different subgraphs and designs a random sampling for the nodes to reduce this overhead. IANS [30] adopts an improved adaptive neighbor sampling to form $k$ subgraphs starting with $k$ central nodes and then defines a subgraph contrastive loss. GNN-LS [31] studies end-to-end learning capability of deep network to realize gradient optimization and samples nodes with an unfixed probability. LoCur [32] extracts subgraphs through combinatorial sampling proportional to localized curvatures with 3-cycles, and demonstrates that the curvatures exhibit small errors in local structures of sparse training graphs. Owing to scalability in terms of layer depth and graph size, these methods exhibit high training efficiency on large-scale graphs [3,6,27-32].

Not extracting only one small subgraph, graph sampling-based methods extract many subgraphs for training; thus, probability models are important for alleviating loss when propagating within and between subgraphs [3,32]. LoCur [32] interpreted the node and edge samplers of GraphSAINT [6] as probability samplings proportional to the localized curvatures without a cycle (or 3-cycle), theoretically providing the error bound of the localized curvatures with 3-cycles on Erdös-Rényi random graphs [33]. Although certain methods may be more affected by scale-free graphs than by mesh grids or Erdös-Rényi random graphs, LoCur [32] exhibits superior performance in various experiments even when considering the computational cost. Our method $\text{HIS}_{\text{GCNs}}$ adopts the framework described in [3,6,32] and focuses on the core-periphery structure unique to scale-free training graphs. The structure formed by preferential attachment [7] helps preserve critical long chains.

## 3 Design principle

We designed a graph sampling-based minibatch learning method called $\text{HIS}_{\text{GCNs}}$ that extracts many subgraphs for information propagation and reduces the variance of node feature aggregation. Our goal differs from graph sampling in network crawling and visualization, which focuses on extracting only one subgraph to represent a certain large-scale graph while minimizing variance in the degree property [11,12]. To mitigate correlation loss between different subgraphs, $\text{HIS}_{\text{GCNs}}$ adopts the loss normalization technique of GraphSAINT [6], which reduces bias toward frequently sampled nodes while incurring minimal overhead in the training time. $\text{HIS}_{\text{GCNs}}$ focuses on preserving important chains in the same subgraph because these chains characterize complex edge correlations, and are helpful in reducing the variance of node feature aggregation.

According to the subgraph-based training framework [3,6,32] and the aforementioned motivation, we designed $HIS_{GCNs}$ based on the following principles:

**P1:** Using a degree threshold to partition the core and periphery, the objective is to maximize the number of edges, each of which connects a core node and periphery node [12].

Nodes with degrees exceeding the threshold were classified as cores, while the rest as peripheries. The maximization method in **P1** can significantly reduce the number of core nodes, which helps preserve the centrum structure of the core in most subgraphs without affecting the diversity of long chains in different subgraphs. Because the node degree is a simple property, the threshold partition incurs slight overhead during training (see **Appendix B.1**).

**P2:** Traversal-based and importance sampling were used to create the periphery of each subgraph. Although the periphery of the large-scale training graph is sparse, the large number of peripheral nodes results in a significant impact of the edges between low-degree nodes on the training. In addition, two connected nodes with few neighbors are likely to influence each other [6]. Thus, without core node interference, traversal-based sampling can preserve an increasing number of longer chains composed entirely of low-degree nodes in the same subgraph. Traditional traversal-based sampling begins with a random seed and iteratively selects one or several neighbors of a traversed node until the desired fraction of nodes is collected [11,34-36]. $HIS_{GCNs}$ adopts these samplings in the periphery construction of each subgraph and designs an importance strategy for sampling the next node from the neighbors of a traversed node objective to reduce node aggregation variance.

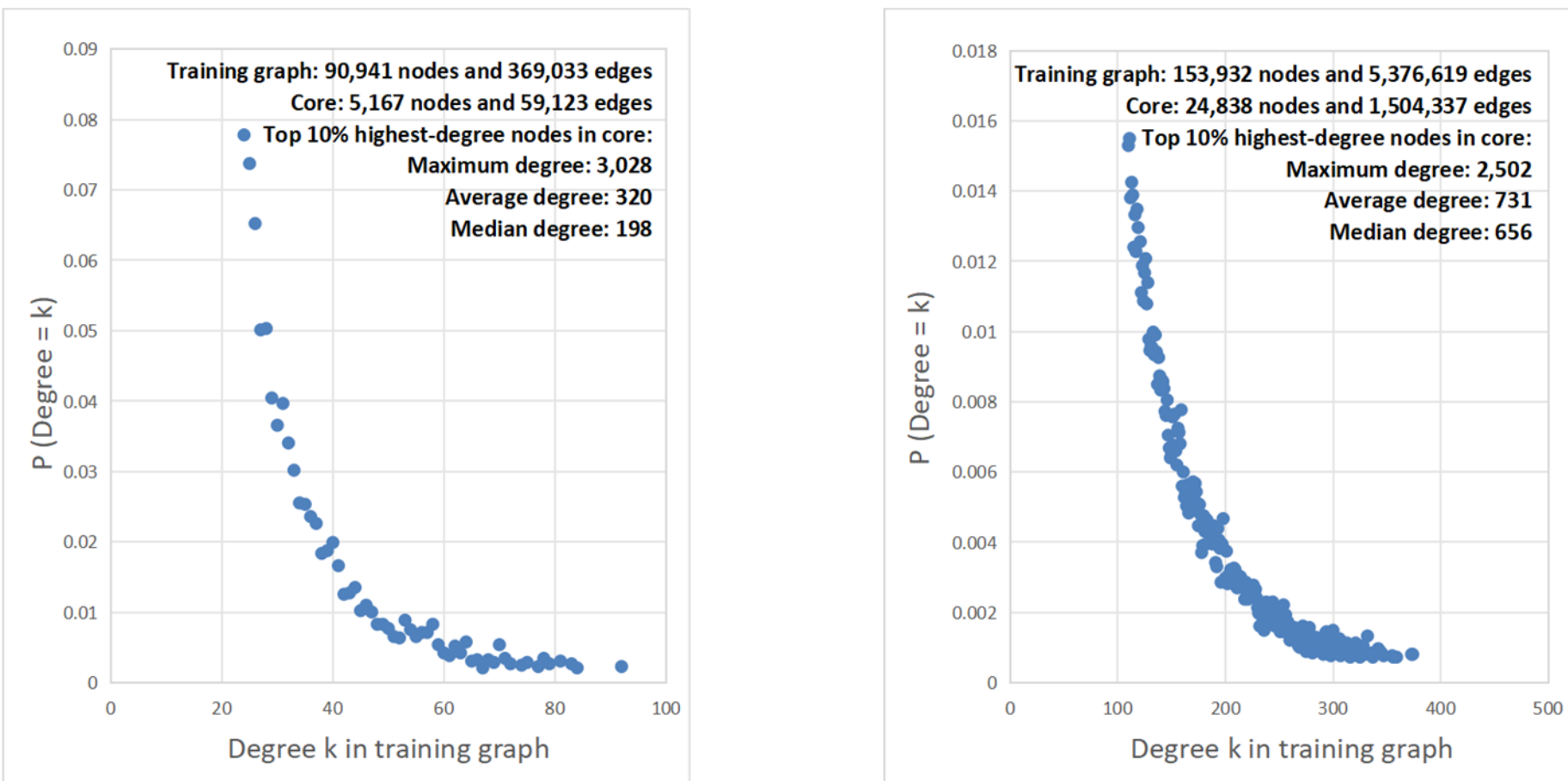

($a$) OGBN-Arxiv [37] ($b$) Reddit [4]

**Figure 1.** Degree distribution of core nodes in training graphs. $P(\text{Degree} = k)$ denotes the ratio of the number of core nodes with degree $k$ to the number of core nodes. The maximum $k$ value of the dots in ($a$)($b$) is equal to the maximum degree of remaining nodes after removing top 10% highest-degree core nodes. The maximum, average and median degrees of the removed core nodes are listed in ($a$)($b$). Centrum nodes are a small number of core nodes with top-highest degrees, and core nodes other than the centrum nodes are defined as marginal core nodes.
The OGBN-Arxiv and Reddit datasets were split into training, validation and test sets (see **Table 1**).

**P3:** With a distribution biased towards high-degree nodes, randomly choose a fraction of core nodes from the neighbors of each periphery node that has been sampled by **P2** and preserve all the chosen core nodes (removing duplicates) in subgraphs.

The periphery nodes are sparsely connected to each other but densely connected to the core nodes in a scale-free graph [10,12], such that most periphery nodes have at least one core neighbor.

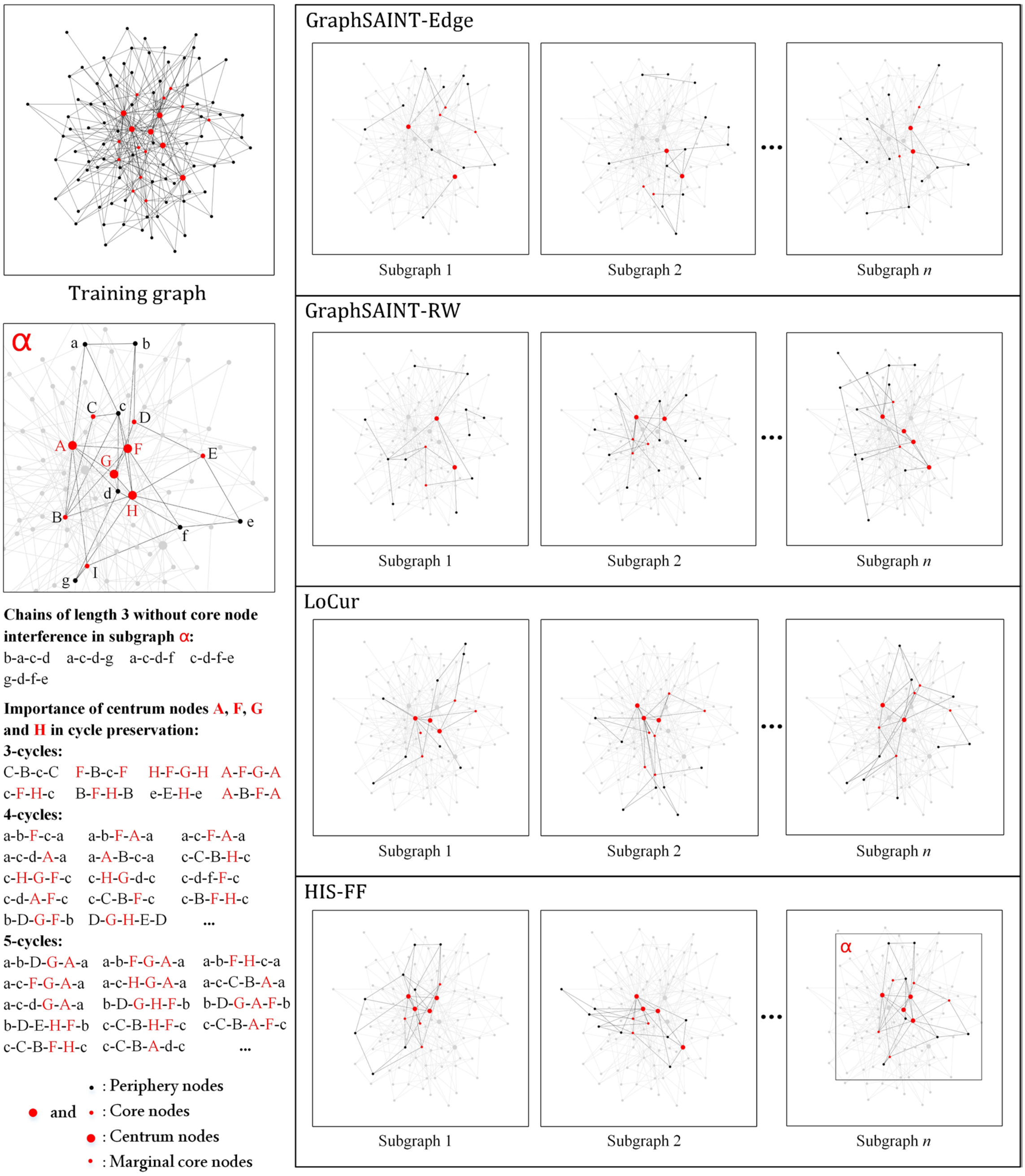

**Figure 2.** Sample examples on a training graph with 100 nodes, in which red and black respectively represent core and periphery nodes. The training graph is generated by Barabasi-Albert (BA) model [7]. Based on **P1**, the degree threshold of the small-scale BA graph is 7, thus nodes with degrees exceeding 7 in the graph are marked in red, while others are marked in black. GraphSAINT-Edge is an edge sampler [6] with $p_{u,v} \propto 1/d_u + 1/d_v$ where $p_{u,v}$ is the probability of sampling edge $(u, v)$ and $d_u, d_v$ denote node degrees. A subgraph is induced by nodes that are end-points of the sampled edges. GraphSAINT-RW is a random walk sampler [6] that first samples $r$ root nodes uniformly at random and then $r$ walkers respectively start from one of the root nodes and each of them goes $h$ hops (the sampler outputs a node induced subgraph). LoCur [32] starts from a set of initial nodes, and then extends to neighboring nodes to maximize the sum of localized curvatures of edges (the localized curvatures are illustrated in **Definition 1**). HIS-FF is one of our $\text{HIS}_{\text{GCNs}}$ samplers that meet **P1** – **P4** and are designed in Section 4. Because only the topology of the BA graph is considered, the 2-norm of the feature vector, i.e., $\|X(v)\|$, for each $v \in \mathcal{V}$ is set to 1. Sample size defined as the number of nodes in the subgraphs is set to 15; however, owing to randomness and the removal of duplicate nodes, the actual sample size usually has a deviation of 1 to 2. Note that partial edges overlap in the graph visualization.

Minibatches are constructed by various subgraphs with a small size; thus, the random choice of the core neighbors in **P3** is helpful in maintaining the diversity of the chains. Removing the core decomposes the scale-free training graph into several branches. Thus, preserving the centrum of the core is important for maintaining the connectivity of the subgraphs. Based on preferential attachment [7], centrum nodes corresponding to a small number of core nodes with the highest degrees are easily connected to periphery nodes. Fig. 1 shows a clear gap between the degrees of the centrum nodes and other marginal core nodes in the large-scale training graphs. The biased random choice in **P3** was executed for each sampled periphery node, making it difficult for the centrum nodes connected by various periphery nodes to fall outside the corresponding subgraph.

**P4:** Preservation of all edges between the sampled nodes in the corresponding subgraph [6].

Based on **P2** and **P3**, sampling from the neighbors of a sampled node can maximize the preservation of connections between two periphery nodes and between the periphery and core nodes. In addition, the core nodes of a scale-free graph, particularly the centrum nodes, are densely interconnected [12]. Thus, **P4** maintains the connectivity of each subgraph.

$\text{HIS}_{\text{GCNs}}$ samples from the periphery to the core, avoiding the high uncertainty of randomly extracting the next node from the neighbors of a centrum node of the core because the number of neighbors in large-scale training graphs is significantly large, as shown in Fig. 1. Owing to the high uncertainty, $\text{HIS}_{\text{GCNs}}$ does not view the centrum node as the starting or ending point of information propagation but only employs it to shorten the length of the propagation path. As shown in Fig. 2, $\text{HIS}_{\text{GCNs}}$ can preserve an increasing number of longer chains in the periphery and connect each subgraph through a few centrum nodes. Owing to various neighbors of a centrum node, the influence of a periphery node (or a marginal core node) that falls within the neighbors of the centrum node is small. However, if two low-degree nodes were attached to the same centrum node, they typically exhibited similar features. As shown in Fig. 3, the centrum nodes play an important role in establishing the correlations between different convolutional layers. Note that the training graph in Fig. 2 has only 100 nodes, resulting in a slight difference of node degrees between the centrum and marginal core; however, this difference is significantly amplified in large-scale graphs, as shown in Fig. 1. In addition, by comparing Figs. 1 and 2, we observe that the ratio of the number of centrum nodes to the total number of nodes in large-scale graphs can be further reduced.

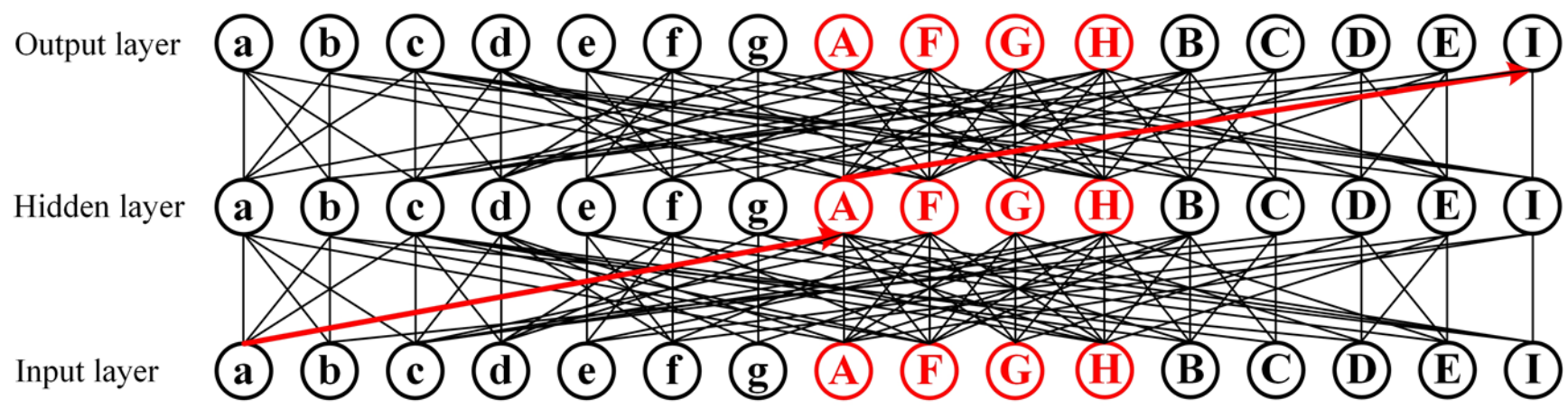

**Figure 3.** Two-layer GCN on subgraph α in left side of Fig. 2.
Centrum node A shortens the path length between periphery node a and marginal core node I, establishing correlations between different convolutional layers.

**Definition 1.** (Localized Curvature with 3-cycles [32]) $d_x$ and $d_y$ denote the degrees of nodes $x$ and $y$, respectively. $\Delta_{\#}(x, y)$ represents the number of triangles (3-cycles) including edge $(x, y)$. $d_x \wedge d_y$ and $d_x \vee d_y$ denote $\min[d_x, d_y]$ and $\max[d_x, d_y]$, respectively. $(\cdot)_+$ is defined as $\max[\cdot, 0]$. Subsequently, localized curvature $\kappa_{xy}$ for edge $(x, y)$ has the following lower bound:

$$\kappa_{xy} \geq -\left(1 - \frac{1}{d_x} - \frac{1}{d_y} - \frac{\Delta_\#(x,y)}{d_x \wedge d_y}\right)_+ - \left(1 - \frac{1}{d_x} - \frac{1}{d_y} - \frac{\Delta_\#(x,y)}{d_x \vee d_y}\right)_+ + \frac{\Delta_\#(x,y)}{d_x \vee d_y} \quad (3)$$

The lower bound of $\kappa_{xy}$ in Eq. (3) can be considered as the localized curvature for the Ollivier-Ricci curvature [38,39], which considers random walk-based probability measures with a Markov chain and 1-Wasserstein transportation distance (see **Appendix A**). The large curvature $\kappa_{xy}$ indicates the small distance between the probability measures for two nodes $x$ and $y$.

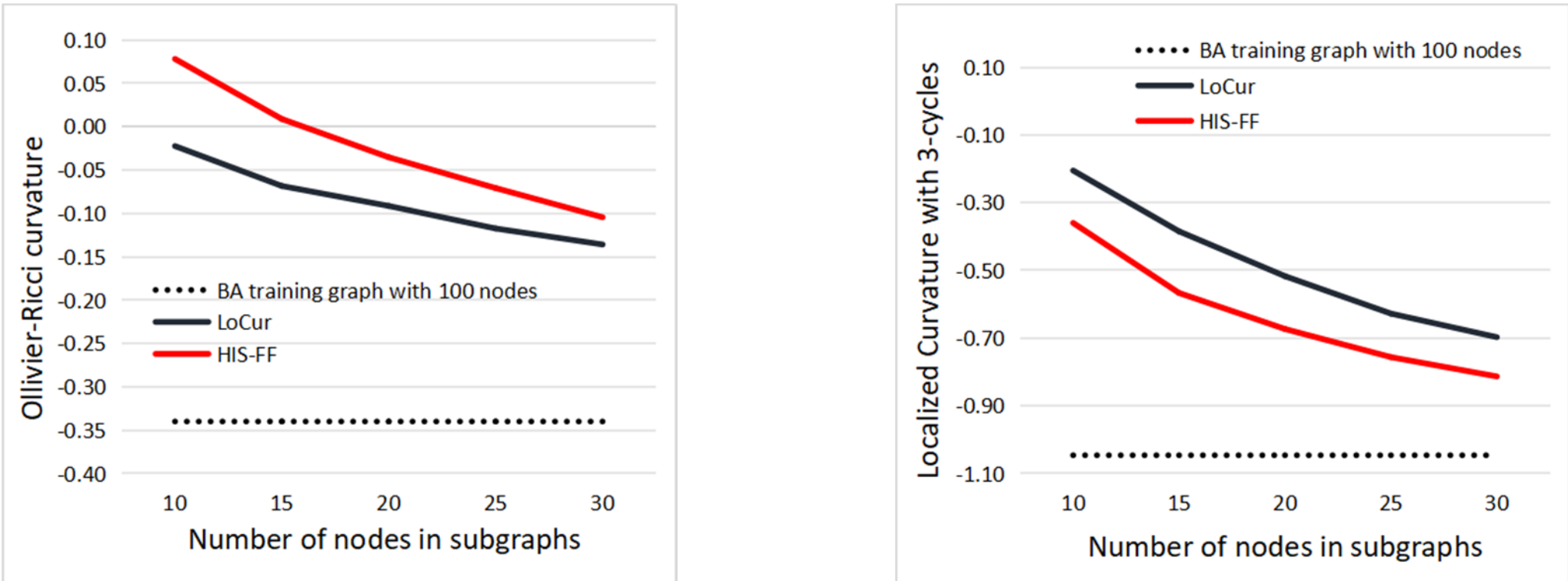

(*a*) Exact curvature *vs.* sample size (*b*) Localized curvature with 3-cycles *vs.* sample size

**Figure 4.** Comparison of Ollivier-Ricci and localized curvatures using the BA training graph with 100 nodes in Fig. 2. The curvatures of LoCur and HIS-FF were evaluated using the average value of edge curvatures of 500 subgraphs for each sample size (i.e., the number of nodes in subgraphs). The curvature of an edge in a subgraph refers to its curvature on the training graph, not on the subgraph. Subgraph sampling that maximizes the edge curvatures amounts to minimizing the distributional difference between the training graph and subgraph [32].

LoCur samples subgraphs to maximize the localized curvature [32]. According to Eq. (3), the lower bound of $\kappa_{xy}$ increases as $d_x$, $d_y$ decrease and $\Delta_\#(x,y)$ increases. Thus, high curvature edges appear in two instances: one is the edges connecting two low-degree nodes, and the other is the edges that are included in various cycles with small length. **P1** – **P4** makes the $\text{HIS}_{\text{GCNs}}$ satisfy the following conditions: preserving the peripheral long chains without core node interference; preserving the core centrum nodes in most subgraphs; and preserving all edges between the sampled nodes. Peripheral long chains are composed of edges that connect low-degree nodes. In scale-free graphs, periphery nodes are densely connected to core nodes (particularly, a few centrum nodes), and the core nodes are densely interconnected [12]; that is, the edges connecting two low-degree periphery nodes are prone to falling into cycles passing through at least one centrum node. In addition, the dense centrum included various cycles with small length. On the left side of Fig. 2, we list two types of chain instances: one represents deep information propagation along low-degree nodes, and the other demonstrates the role of centrum nodes in preserving cycles that shorten the propagation distance between the periphery or marginal core nodes. The experiments in Fig. 4 show that $\text{HIS}_{\text{GCNs}}$ and LoCur have the same goal of maximizing the curvatures.

We designed $\text{HIS}_{\text{GCNs}}$ with the following novelty:

- Different from network crawling and visualization that retain global structures of an original graph to a unique subgraph, $\text{HIS}_{\text{GCNs}}$ aims to preserve local structures of deep propagation chains to the same subgraph that has a smaller size, and samples many subgraphs with the assistance of the core-periphery partition to avoid the loss of critical information.

The Ollivier-Ricci curvature was calculated by open sourced code (see Appendix A for parameter setting) [39]: *https://github.com/saibalmars/GraphRicciCurvature?tab=readme-ov-file*

- We designed two subgraph samplers, HIS-FF and HIS-RW, for $HIS_{GCNs}$ to preserve critical chains and reduce node aggregation variance.
- We theoretically analyzed the different effects of core and periphery nodes on node aggregation variance and then experimentally analyzed the superior performance of $HIS_{GCNs}$ in reducing variance on multi-layer GCNs with non-linear activations.
- We analyzed the favorable and unfavorable roles of centrum nodes in node classification tasks based on the unique structure of scale-free graphs and analyzed the superior performance of $HIS_{GCNs}$ in chain preservation, curvature maximization, convergence and accuracy. Specifically, our method improves the average F1-micro score on the test set by over $1\%$ across six selected datasets, outperforming other subgraph samplers.
- $HIS_{GCNs}$ maximizes the edge curvatures without counting the number of triangles, further improving the time efficiency. Using the core-periphery structures, the sampler HIS-RW of $HIS_{GCNs}$ can reduce the sampling time consumption by more than $55\%$ on large-scale training graphs compared with GraphSAINT-RW.

# 4 Proposed method: $HIS_{GCNs}$

## 4.1 Core-periphery partition

Based on **P1**, we adopted the method in [12] that defines a degree threshold $d_{th}$ to partition the core and periphery in the scale-free training graph $G = (\mathcal{V}, \mathcal{E})$ with node set $\mathcal{V}$ and edge set $\mathcal{E}$.

$$d_{th} = \underset{d}{\operatorname{argmax}} |\{(u, v) \in \mathcal{E} \mid u, v \in \mathcal{V}, d_u > d, d_v \leq d\}| \tag{4}$$

where $d_u$ and $d_v$ respectively denote the degrees of nodes $u$ and $v$ in graph $G$, and $|\cdot|$ is defined as the cardinality of a set. Then, $G$ is partitioned into a small and dense core $G_{cor} = (\mathcal{V}_{cor}, \mathcal{E}_{cor})$, a large and sparse periphery $G_{per} = (\mathcal{V}_{per}, \mathcal{E}_{per})$, and a vertical edge set $\mathcal{E}_{ver}$:

$$\mathcal{V}_{cor} = \{ v \in \mathcal{V} \mid d_v > d_{th} \}, \quad \mathcal{E}_{cor} = \{ (u, v) \in \mathcal{E} \mid u, v \in \mathcal{V}_{cor} \} \tag{5}$$

$$\mathcal{V}_{per} = \{ v \in \mathcal{V} \mid d_v \leq d_{th} \}, \quad \mathcal{E}_{per} = \{ (u, v) \in \mathcal{E} \mid u, v \in \mathcal{V}_{per} \} \tag{6}$$

$$\mathcal{E}_{ver} = \{ (u, v) \in \mathcal{E} \mid u \in \mathcal{V}_{cor}, v \in \mathcal{V}_{per} \} \tag{7}$$

In scale-free graph $G$, the probability of randomly selecting a $k$-degree node decays as a power law $P(k) \propto k^{-\tau}$, where $\tau$ is the degree exponent [7]. Specifically, various low-degree nodes exist in $G_{per}$. Thus, we can increase the degree threshold $d_{th}$ from 1 with a step length of 1 until Eq. (4) is satisfied [12]. There are two advantages of the degree threshold partition: it minimizes the number of nodes in $G_{cor}$, and the overhead is small (see **Appendix B.1**).

## 4.2 Hierarchical importance sampling

We derived samplers to reduce the variance of the node information aggregation in Eq. (2). Based on **P2**, during the process of traversing neighbors in the periphery, we define $p(u \mid v)$ as the probability of sampling $u \in \mathcal{V}_{per}$ from the neighbors of a particular traversed periphery node $v$. In addition, based on **P3**, during the process of choosing core neighbors, we define $q(u' \mid v)$ as the probability of sampling $u' \in \mathcal{V}_{cor}$ from the neighbors of a certain sampled periphery node $v$.

The defined probabilities meet the following constraints:

$$\sum_{u \in \mathcal{V}_{per}} p(u \mid v) = 1, \qquad \sum_{u' \in \mathcal{V}_{cor}} q(u' \mid v) = 1 \tag{8}$$

To derive $p(u \mid v)$, we employ a theoretical method under the assumption of layer independence. This is similar to the independent treatment of layers in prior studies [6,25]. To obtain $q(u' \mid v)$, we rely on the empirical analysis on centrum nodes in Section 3, that is, the centrum nodes play an important role in preserving cycles that are entangled with peripheral long chains and establishing complex correlations across different layers. The effectiveness of the variance reduction in the empirical analysis is validated in Section 5.3.

Under the assumption of layer independence and based on Eq. (2), we transform the embedding of periphery node $v$ at layer $l+1$ as follows:

$$H^{(l+1)}(v) = \left( \sum_{u \in \mathcal{V}_{per}} \hat{A}_{v,u} H^{(l)}(u) + \sum_{u' \in \mathcal{V}_{cor}} \hat{A}_{v,u'} H^{(l)}(u') \right) W^{(l)} \tag{9}$$

The vast majority of nodes in the scale-free training graph were allocated to the periphery; thus, Eq. (9) considers only the embedding of nodes in the periphery.

Based on Eq. (8), we derive two expectation forms:

$$\mu_{per}(v) = \mathbb{E}_{p(u|v)}\left[\frac{1}{p(u \mid v)} \hat{A}_{v,u} H^{(l)}(u)\right] = \sum_{u \in \mathcal{V}_{per}} \left[\frac{1}{p(u \mid v)} \hat{A}_{v,u} H^{(l)}(u) \cdot p(u \mid v)\right] \tag{10}$$

$$\mu_{cor}(v) = \mathbb{E}_{q(u'|v)}\left[\frac{1}{q(u' \mid v)} \hat{A}_{v,u'} H^{(l)}(u')\right] = \sum_{u' \in \mathcal{V}_{cor}} \left[\frac{1}{q(u' \mid v)} \hat{A}_{v,u'} H^{(l)}(u') \cdot q(u' \mid v)\right] \tag{11}$$

Based on Eqs. (10) and (11), Eq.(9) can be rewritten as follows:

$$H^{(l+1)}(v) = \left(\mu_{per}(v) + \mu_{cor}(v)\right) W^{(l)} \tag{12}$$

We evaluate the expectations $\mu_{per}(v)$ and $\mu_{cor}(v)$ using Monte Carlo sampling [23]:

$$\hat{\mu}_{per}(v) = \frac{1}{n} \sum_{i=1}^{n} \frac{1}{p(u_i \mid v)} \hat{A}_{v,u_i} H^{(l)}(u_i), \qquad u_i \sim p(u \mid v) \tag{13}$$

$$\hat{\mu}_{cor}(v) = \frac{1}{n} \sum_{i=1}^{n} \frac{1}{q(u_i' \mid v)} \hat{A}_{v,u_i'} H^{(l)}(u_i'), \qquad u_i' \sim q(u' \mid v) \tag{14}$$

Following the approach in [6], we first assume that $H^{(l)}(u_i)$ is one dimensional (i.e., a scalar), $i = 1,2,\cdots,n$. We derive the variance of Eq. (13) and calculate $p(u \mid v)$ that minimizes the variance.

$$Var[\hat{\mu}_{\mathrm{per}}(v)] = \mathbb{E}_{p(u|v)}\left[\left(\frac{1}{n} \sum_{i=1}^{n} \frac{1}{p(u_i \mid v)} \hat{A}_{v,u_i} H^{(l)}(u_i)\right)^2\right] - \left[\mu_{per}(v)\right]^2 \tag{15}$$

Based on iid samples $u_1, u_2, \cdots, u_n \sim p(u \mid v)$, Eq. (8) and Eq. (10), we obtain:

$$\begin{aligned}
\mathbb{E}_{p(u|v)}\left[\left(\frac{1}{n} \sum_{i=1}^{n} \frac{1}{p(u_i \mid v)} \hat{A}_{v,u_i} H^{(l)}(u_i)\right)^2\right] &= \mathbb{E}_{p(u|v)}\left[\left(\frac{1}{n} \sum_{i=1}^{n} \frac{1}{p(u_i \mid v)} \hat{A}_{v,u_i} H^{(l)}(u_i)\right)\left(\frac{1}{n} \sum_{j=1}^{n} \frac{1}{p(u_j \mid v)} \hat{A}_{v,u_j} H^{(l)}(u_j)\right)\right] \\
&= \mathbb{E}_{p(u|v)}\left[\frac{1}{n^2} \sum_{i=1}^{n} \sum_{j=1}^{n} \frac{1}{p(u_i \mid v)} \hat{A}_{v,u_i} H^{(l)}(u_i) \frac{1}{p(u_j \mid v)} \hat{A}_{v,u_j} H^{(l)}(u_j)\right] \\
&= \frac{1}{n^2} \sum_{i=1}^{n} \mathbb{E}_{p(u_i|v)}\left[\frac{1}{p(u_i \mid v)} \hat{A}_{v,u_i} H^{(l)}(u_i)\right]^2 \\
&+ \frac{2}{n^2} \sum_{i=1}^{n-1} \sum_{j=i+1}^{n} \left\{\mathbb{E}_{p(u_i|v)}\left[\frac{1}{p(u_i \mid v)} \hat{A}_{v,u_i} H^{(l)}(u_i)\right] \cdot \mathbb{E}_{p(u_j|v)}\left[\frac{1}{p(u_j \mid v)} \hat{A}_{v,u_j} H^{(l)}(u_j)\right]\right\} \\
&= \frac{1}{n} \mathbb{E}_{p(u|v)}\left[\frac{1}{p(u \mid v)} \hat{A}_{v,u} H^{(l)}(u)\right]^2 + \left(1 - \frac{1}{n}\right)\left\{\mathbb{E}_{p(u|v)}\left[\frac{1}{p(u \mid v)} \hat{A}_{v,u} H^{(l)}(u)\right]\right\}^2 \\
&= \frac{1}{n} \sum_{u \in \mathcal{V}_{per}} \left\{\left[\frac{1}{p(u \mid v)} \hat{A}_{v,u} H^{(l)}(u)\right]^2 \cdot p(u \mid v)\right\} + \left(1 - \frac{1}{n}\right)\left[\mu_{per}(v)\right]^2 \\
&= \frac{1}{n} \sum_{u \in \mathcal{V}_{per}} \left[\frac{1}{p(u \mid v)} \hat{A}_{v,u}^2 \left|H^{(l)}(u)\right|^2\right] + \left(1 - \frac{1}{n}\right)\left[\mu_{per}(v)\right]^2
\end{aligned}$$

Thus, we transform Eq. (15) to the following form:

$$Var[\hat{\mu}_{per}(v)] = \frac{1}{n} \sum_{u \in \mathcal{V}_{per}} \left[\frac{1}{p(u \mid v)} \hat{A}_{v,u}^{2} \left|H^{(l)}(u)\right|^{2}\right] - \frac{1}{n}\left[\mu_{per}(v)\right]^{2} \tag{16}$$

Based on Eq.(10), Eq. (16) can be rewritten as follows:

$$Var[\hat{\mu}_{per}(v)] = \frac{1}{n} \sum_{u \in \mathcal{V}_{per}} \left[\frac{1}{p(u \mid v)} \hat{A}_{v,u}^{2} \left|H^{(l)}(u)\right|^{2}\right] - \frac{1}{n}\left[\sum_{u \in \mathcal{V}_{per}} \hat{A}_{v,u} H^{(l)}(u)\right]^{2} \tag{17}$$

Eq. (8) shows that $\sum_{u \in \mathcal{V}_{per}} p(u \mid v) = 1$. By Cauchy-Schwarz inequality:

$$\sum_{u \in \mathcal{V}_{per}} \left[\frac{1}{p(u \mid v)} \hat{A}_{v,u}^{2} \left|H^{(l)}(u)\right|^{2}\right] \sum_{u \in \mathcal{V}_{per}} p(u \mid v) = \sum_{u \in \mathcal{V}_{per}} \left[\frac{\hat{A}_{v,u} \left|H^{(l)}(u)\right|}{\sqrt{p(u \mid v)}}\right]^{2} \sum_{u \in \mathcal{V}_{per}} \left(\sqrt{p(u \mid v)}\right)^{2} \geq \left[\sum_{u \in \mathcal{V}_{per}} \hat{A}_{v,u} \left|H^{(l)}(u)\right|\right]^{2}$$

The equality is achieved when $\frac{\hat{A}_{v,u}|H^{(l)}(u)|}{\sqrt{p(u|v)}} \propto \sqrt{p(u \mid v)}$, i.e., variance is minimized when

$$p(u \mid v) \propto \hat{A}_{v,u} \left|H^{(l)}(u)\right| \tag{18}$$

For the multi-dimensional case of $H^{(l)}(u)$, by following the steps described above, it is easy to demonstrate that the optimal sampling probability to minimize the variance is

$$p(u \mid v) \propto \hat{A}_{v,u} \left\|H^{(l)}(u)\right\| \tag{19}$$

where $\|\cdot\|$ denotes 2-norm with $\|(x_1, x_2, \cdots, x_k)\| = \left(\sum_{i=1}^{k} x_i^2\right)^{1/2}$.

As in [25], the initial feature vector $X(u)$ defined in Section 2.2 was used to approximate the hidden feature vector $H^{(l)}(u)$. Thus, Eq. (19) can be simplified as follows:

$$p(u \mid v) \propto \hat{A}_{v,u} \|X(u)\| \tag{20}$$

In the GCNs described in Section 2.2, $\hat{A}_{v,u} = 1/\sqrt{(d_v + 1)(d_u + 1)}$, where $d_v$ and $d_u$ denote the degrees of nodes $v$ and $u$, respectively. As $v$ is a specified node, $\hat{A}_{v,u} \propto 1/\sqrt{d_u + 1}$.

Thus, we rewrite Eq. (20) in the following form:

$$p(u \mid v) \propto \frac{\|X(u)\|}{\sqrt{d_u + 1}} \tag{21}$$

Similar to Eqs. (15) to (21), we derive $q(u' \mid v) \propto \|X(u')\|/\sqrt{d_{u'} + 1}$. This result **conflicts with P3**, because $q(u' \mid v)$ decreases with increasing $d_{u'}$ where $u' \in \mathcal{V}_{cor}$; however, **P3** needs a probability distribution that is biased towards high-degree nodes. Similar to Eq. (17), we derive

$$Var[\hat{\mu}_{\mathrm{cor}}(v)] = \frac{1}{n} \sum_{u' \in \mathcal{V}_{cor}} \left[\frac{1}{q(u' \mid v)} \hat{A}_{v,u'}^{2} \left|H^{(l)}(u')\right|^{2}\right] - \frac{1}{n}\left[\sum_{u' \in \mathcal{V}_{cor}} \hat{A}_{v,u'} H^{(l)}(u')\right]^{2} \tag{22}$$

However, $\hat{A}_{v,u'} = 1/\sqrt{(d_v + 1)(d_{u'} + 1)}$ is significantly small for core node $u'$ in Eq. (22), especially for the centrum node $u'$, because the degree of the core node (especially the centrum node) is significantly higher than that of the periphery node, as shown in Fig. 1. In addition, the percentage of core nodes in large scale-free graphs is low. For example, the ratio of the number of core nodes to the total number of nodes is $5.68\%(\frac{5{,}167}{90{,}941})$ in Fig. 1($a$) and $16.13\%(\frac{24{,}838}{153{,}932})$ in Fig. 1($b$). In particular, $|\mathcal{V}_{cor}| \ll |\mathcal{V}_{per}|$.

Therefore, based on Eqs. (12), (17) and (22), Eq. (22) can be considered as noise in the analysis of variance theory. To derive $q(u' \mid v)$, we adopt the empirical analysis on centrum nodes in Section 3, that is, preserving the centrum in most subgraphs is important for $\mathrm{HIS_{GCNs}}$. Thus, we empirically set $q(u' \mid v) \propto \hat{A}_{v,u'} \|X(u')\|(d_{u'} + 1)$, namely

$$q(u' \mid v) \propto \|X(u')\|\sqrt{d_{u'} + 1} \tag{23}$$

It is well-known that $q(u' \mid v)$ in Eq. (23) is biased towards high-degree nodes $u'$.

## 4.3 Pseudo codes and flowchart

Preprocessing, especially the core-periphery partition on the training graph, is executed before subgraph sampling; it runs only once and requires minimal time. To obtain $n$ subgraphs, $\text{HIS}_{\text{GCNs}}$ adopts a traversal-based method that samples nodes in the periphery $G_{per} = (\mathcal{V}_{per}, \mathcal{E}_{per})$ without core node interference, and preserves a fraction of core neighbors for each newly traversed periphery node, until the number of sampled periphery and core nodes satisfies the desired sample size. The subgraphs preserve all the edges between the sampled nodes. At each iteration, for a traversed periphery node $v$, $\text{HIS}_{\text{GCNs}}$ samples its periphery neighbors with $u_1, u_2, \cdots, u_s \sim p(u \mid v)$ and its core neighbors with $u'_1, u'_2, \cdots, u'_t \sim q(u' \mid v)$, where $s$ is determined by the chosen traversal-based method and $t = \langle \gamma \cdot |\{u' \in \mathcal{V}_{cor} | (u', v) \in \mathcal{E}_{ver}\}| \rangle$. Note that $\mathcal{E}_{ver}$ is defined in Eq. (7) and $\langle \cdot \rangle$ rounds a scalar to the nearest integer. Parameter $\gamma \in (0,1]$ decreases with growth of the scale of the training graph, because the difference of node degrees between centrum and marginal core is amplified on large-scale graphs. We experimentally set $\gamma = 0.4$ for training graphs containing more than $100{,}000$ nodes; otherwise $\gamma = 1$. In this section, $\text{HIS}_{\text{GCNs}}$ updates two traversal-based methods, namely Forest Fire (FF) [35] and GraphSAINT-RW [6], and their pseudocodes are listed in Algorithms 1 and 2.

---

**Algorithm 1:** HIS-FF

---

**Input:** Scale-free training graph $G = (\mathcal{V}, \mathcal{E})$, core $G_{cor} = (\mathcal{V}_{cor}, \mathcal{E}_{cor})$, periphery $G_{per} = (\mathcal{V}_{per}, \mathcal{E}_{per})$, vertical edge set $\mathcal{E}_{ver}$, parameter $\gamma$, sample size $\hat{n}$.
**Output:** Subgraph $G_{sub} = (\mathcal{V}_{sub}, \mathcal{E}_{sub})$.

1: Initialize an empty FIFO (First-In First-Out) queue $Q$, $\mathcal{V}_{sub} \leftarrow \emptyset$.
2: **While** $|\mathcal{V}_{sub}| < \hat{n}$ **do**
3: **If** $Q$ is empty, **then** uniformly at random choose a seed node $w \notin \mathcal{V}_{sub}$ from $\mathcal{V}_{per}$ and add $w$ to $Q$.
4: Extract and delete node $v$ from $Q$, and update $\mathcal{V}_{sub} \leftarrow \mathcal{V}_{sub} \cup \{v\}$.
5: Sample $t = \langle \gamma \cdot |\mathcal{N}_{cor}(v)| \rangle$ nodes $u'_1, u'_2, \cdots, u'_t \sim q(u' \mid v) = \|X(u')\|\sqrt{d_{u'} + 1} / \sum_{u' \in \mathcal{N}_{cor}(v)} \|X(u')\|\sqrt{d_{u'} + 1}$ from $\mathcal{N}_{cor}(v) = \{u' \in \mathcal{V}_{cor} | (u', v) \in \mathcal{E}_{ver}\}$, and update $\mathcal{V}_{sub} \leftarrow \mathcal{V}_{sub} \cup \{u'_1, u'_2, \cdots, u'_t\}$.
6: Generate a random number $s \sim$ Geometric distribution with mean $(1-p)^{-1}$, where $p$ is experimentally set to 0.5. **If** $|\mathcal{V}_{sub}| + |Q| < \hat{n}$, **then** sample $s$ nodes $u_1, u_2, \cdots, u_s \sim p(u \mid v) = (\|X(u)\|/\sqrt{d_u + 1}) / \sum_{u \in \mathcal{N}_{per}(v)} (\|X(u)\|/\sqrt{d_u + 1})$ from $\mathcal{N}_{per}(v) = \{u \in \mathcal{V}_{per} | (u, v) \in \mathcal{E}_{per}, u \notin \mathcal{V}_{sub}, u \notin Q\}$, and add $u_1, u_2, \cdots, u_s$ to $Q$.
7: **End while** #Annotation: $|\mathcal{V}_{sub} \cup \{u'_1, u'_2, \cdots, u'_t\}| \approx |\mathcal{V}_{sub}|$ in line 5, since many periphery nodes are connected to the same centrum node.
8: $G_{sub} = (\mathcal{V}_{sub}, \mathcal{E}_{sub}) \leftarrow$ Node induced subgraph of $G$ from $\mathcal{V}_{sub}$.

---

**Algorithm 2:** HIS-RW

---

**Input:** Scale-free training graph $G = (\mathcal{V}, \mathcal{E})$, core $G_{cor} = (\mathcal{V}_{cor}, \mathcal{E}_{cor})$, periphery $G_{per} = (\mathcal{V}_{per}, \mathcal{E}_{per})$, vertical edge set $\mathcal{E}_{ver}$, parameter $\gamma$, sample size $\hat{n}$, walk length $h$.
**Output:** Subgraph $G_{sub} = (\mathcal{V}_{sub}, \mathcal{E}_{sub})$.

1: Initialize $\mathcal{V}_{sub} \leftarrow \emptyset$.
2: **While** $|\mathcal{V}_{sub}| < \hat{n}$ **do**
3: Uniformly at random choose a seed node $v$ from $\mathcal{V}_{per}$.
4: **For** $d = 1$ to $h + 1$ **do**
5: **If** $v \notin \mathcal{V}_{sub}$, **then** sample $t = \langle \gamma \cdot |\mathcal{N}_{cor}(v)| \rangle$ nodes $u'_1, u'_2, \cdots, u'_t \sim q(u' \mid v) = \|X(u')\|\sqrt{d_{u'} + 1} / \sum_{u' \in \mathcal{N}_{cor}(v)} \|X(u')\|\sqrt{d_{u'} + 1}$ from $\mathcal{N}_{cor}(v) = \{u' \in \mathcal{V}_{cor} | (u', v) \in \mathcal{E}_{ver}\}$, update $\mathcal{V}_{sub} \leftarrow \mathcal{V}_{sub} \cup \{u'_1, u'_2, \cdots, u'_t\}$, and update $\mathcal{V}_{sub} \leftarrow \mathcal{V}_{sub} \cup \{v\}$.
6: **If** $d < h + 1$, **then** sample a node $u_1 \sim p(u \mid v) = (\|X(u)\|/\sqrt{d_u + 1}) / \sum_{u \in \mathcal{N}_{per}(v)} (\|X(u)\|/\sqrt{d_u + 1})$ from $\mathcal{N}_{per}(v) = \{u \in \mathcal{V}_{per} | (u, v) \in \mathcal{E}_{per}\}$, and update $v \leftarrow u_1$.
7: **End For**
8: **End while**
9: $G_{sub} = (\mathcal{V}_{sub}, \mathcal{E}_{sub}) \leftarrow$ Node induced subgraph of $G$ from $\mathcal{V}_{sub}$.

---

An overall flowchart of HIS-FF and HIS-RW is shown in Fig. 5.

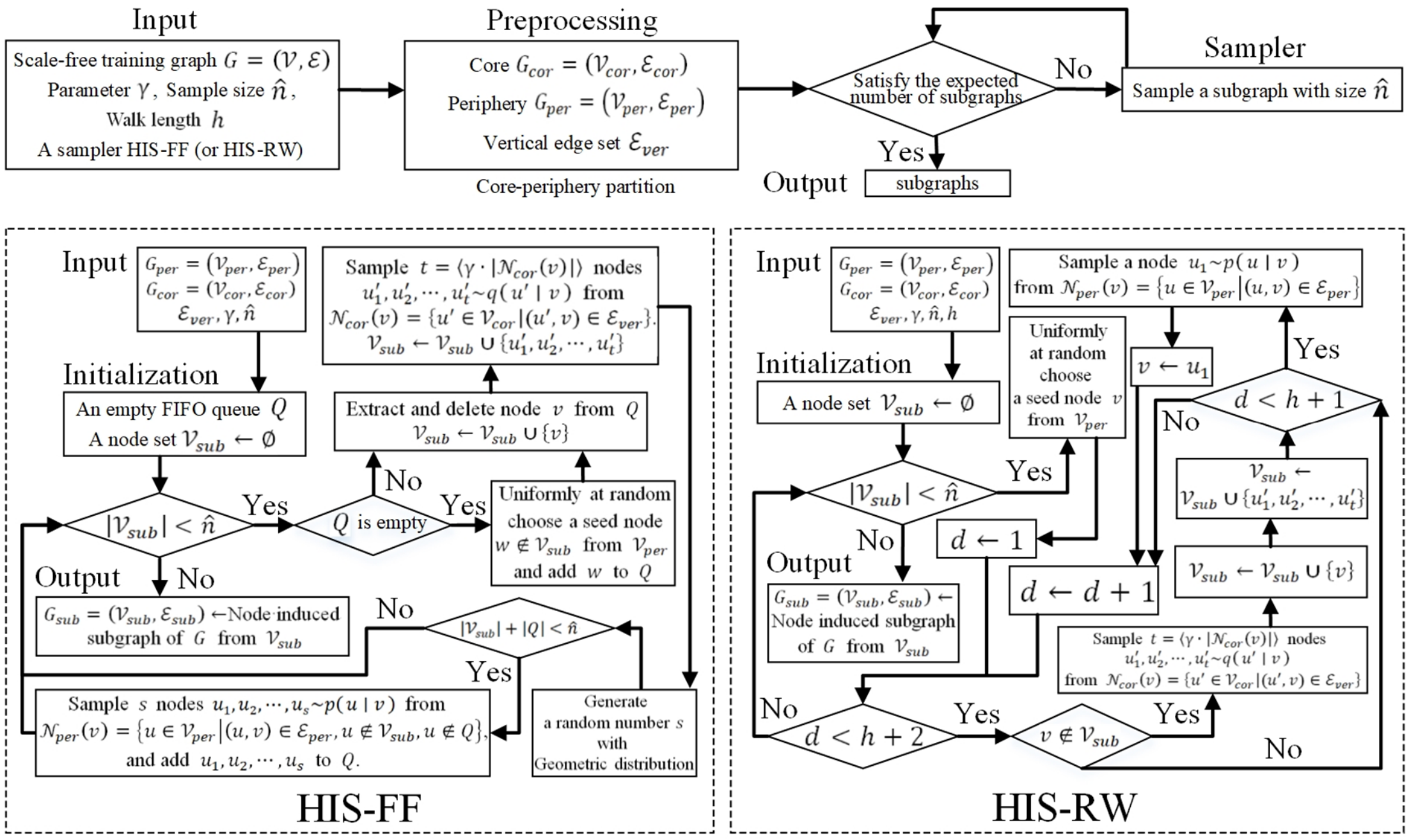

**Figure 5.** Overall flowchart of HIS-FF and HIS-RW.

# 5 Experiments

In this section, we validate the effectiveness of HIS-FF and HIS-RW on chain/cycle preservation and variance reduction, and compare our method $\mathrm{HIS}_{\mathrm{GCNs}}$ with several baseline algorithms on classic graph datasets for semi-supervised node classification tasks. The training algorithm of our method $\mathrm{HIS}_{\mathrm{GCNs}}$ can be found in **Appendix C**. Experiments were conducted on a desktop computer equipped with a GPU 4070 and a 3.50 GHz 13th Gen Intel(R) Core(TM) i5-13600KF CPU, with 32GB of RAM and 16GB of GPU memory, and implemented in PyTorch and Python 3.

**Table 1.** Dataset statistics ("m" stands for multi-class classification, and "s" for single-class). In the training graphs, a few isolated nodes with degree zero have been removed.

| Dataset | Nodes | Edges | Features | Classes | Train/ Validation/ Test | Training graph | |
|---|---|---|---|---|---|---|---|
| | | | | | | Nodes | Edges |
| CiteSeer | 3,327 | 4,732 | 500 | 6 (s) | 0.54/0.15/0.31 | 1,812 | 1,351 |
| Pubmed | 19,717 | 44,338 | 500 | 3 (s) | 0.92/0.03/0.05 | 18,217 | 37,900 |
| PPI-Large | 56,944 | 818,716 | 50 | 121 (m) | 0.79/0.11/0.10 | 44,906 | 633,198 |
| Ogbn-arxiv | 169,343 | 1,166,243 | 128 | 40 (m) | 0.54/0.18/0.28 | 90,941 | 369,033 |
| Reddit | 232,965 | 11,606,919 | 602 | 41 (s) | 0.66/0.10/0.24 | 153,932 | 5,376,619 |
| Ogbn-products | 2,449,026 | 61,859,140 | 100 | 47 (s) | 0.08/0.02/0.90 | 196,615 | 5,451,633 |

## 5.1 Graph datasets

We chose six scale-free datasets for the node classification tasks listed in Table 1. The datasets were divided into training, validation and test sets. CiteSeer [40] and Pubmed [41] are two classical small-scale datasets split using the method in [24]. PPI-Large, an abbreviation of PPI (large version) [6], is a Protein-Protein Interaction network, Reddit [4] is a social network, OGBN-Arxiv [37] is a citation network, and OGBN-Products [37] is a product information network. The training, validation and test sets of PPI-Large, Reddit, OGBN-Arxiv and OGBN-Products were split into datasets. In addition, some graphs from the Stanford large network dataset collection [42] were used to analyze the exact Ollivier-Ricci curvature and the core-periphery partition.

Note that the sample examples in Fig. 2 and the curvature comparisons in Fig. 4 and Table 2 consider only the topological structure of the subgraphs; thus, $\|X(u)\|$ in Eq. (21) and $\|X(u')\|$ in Eq. (23) of HIS-FF and HIS-RW are all set to 1 for $\forall u \in \mathcal{V}_{per}, u' \in \mathcal{V}_{cor}$ in these experiments.

### 5.2 Chain preservation and curvature maximization

Compared to the supporting role of centrum nodes in shortening the path length between two low-degree nodes, information propagation along consecutive low-degree nodes in a long chain is more effective. Let $\mathcal{P}_G(k)$ denote the set of all chains composed of four nodes with degrees not more than $k$ in the training graph $G$. Define $\mathbf{1}(\tau \subseteq G_i)$ is 1 if chain $\tau$ exists in subgraph $G_i$, otherwise 0, and let $x \vee y$ denote $\max[x, y]$. Subsequently, we employ $r_k^n$ defined in Eq. (24) to measure the preservation rate of the $\mathcal{P}_G(k)$ chains in subgraphs $G_i$ $(i = 1,2,\cdots,n)$.

$$r_k^n = \frac{1}{|\mathcal{P}_G(k)|} \sum_{\tau \in \mathcal{P}_G(k)} \bigvee_{i=1}^{n} \mathbf{1}(\tau \subseteq G_i) \tag{24}$$

Eq. (24) confirms that chain $\tau$ is preserved provided it falls into one of the $n$ subgraphs, as $\text{HIS}_{\text{GCNs}}$ adopts the loss normalization technique of GraphSAINT that eliminates bias to nodes more frequently sampled in different subgraphs. As shown in Fig. 6, we compare HIS-FF and HIS-RW with the existing graph samplers introduced in Section 2.3.3 and Fig. 2, which validates the effectiveness of preserving long chains composed entirely of low-degree nodes.

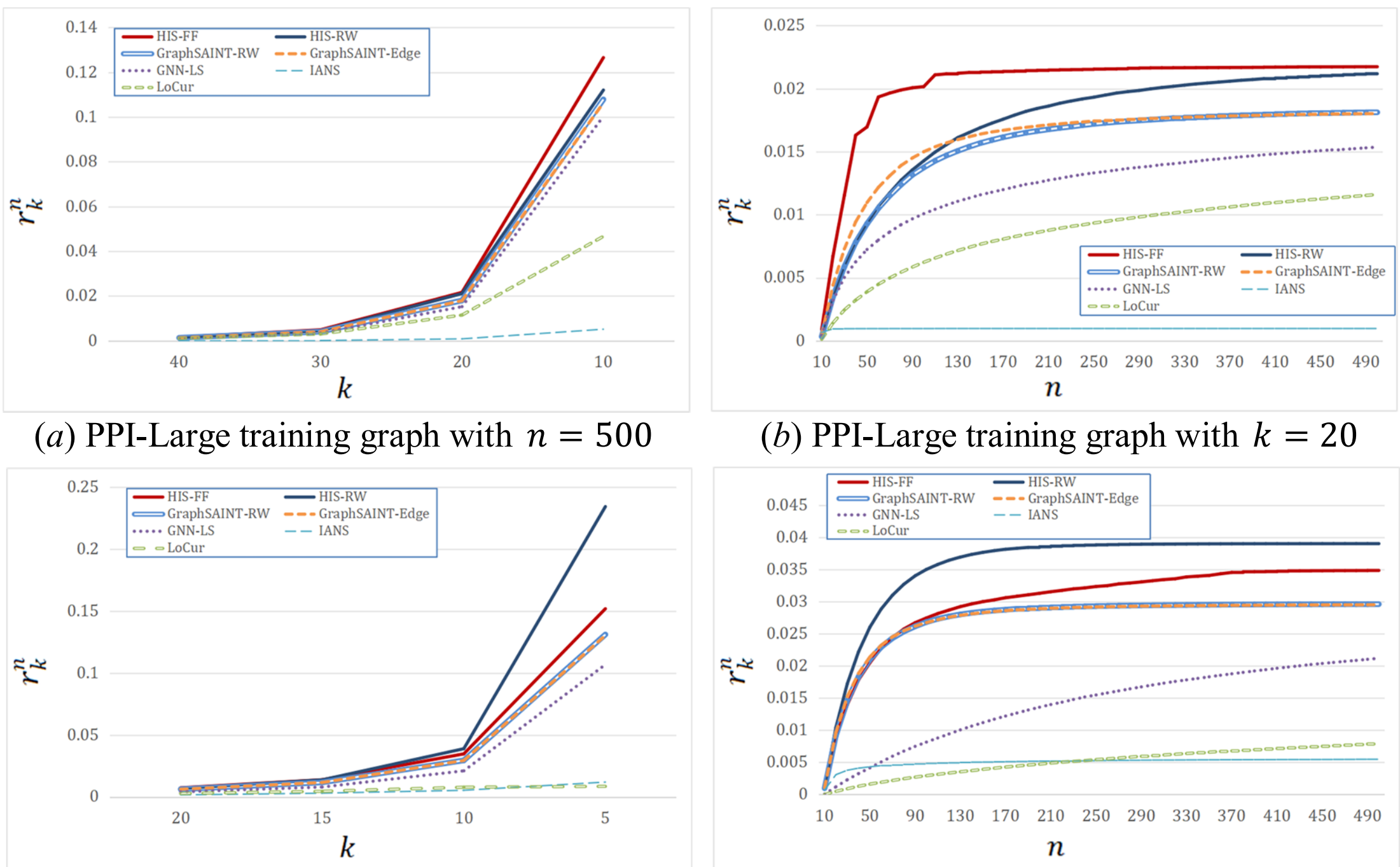

(*a*) PPI-Large training graph with $n = 500$ (*b*) PPI-Large training graph with $k = 20$

(*c*) Ogbn-arxiv training graph with $n = 500$ (*d*) Ogbn-arxiv training graph with $k = 10$

**Figure 6.** Comparison of preservation rate of chains along low-degree nodes. (*a*) $r_k^n$ *vs.* $k$ with $n = 500$ in PPI-Large training graph. (*b*) $r_k^n$ *vs.* $n$ with $k = 20$ in PPI-Large training graph. (*c*) $r_k^n$ *vs.* $k$ with $n = 500$ in Ogbn-arxiv training graph. (*d*) $r_k^n$ *vs.* $n$ with $k = 10$ in Ogbn-arxiv training graph. The measure $r_k^n$ is defined in Eq. (24).

When paths traverse edges with high curvatures, the distance between the feature representations of the starting and ending nodes of a path is reduced [32]. Therefore, it is necessary to maximize the curvatures of a subgraph. In 1-hop local structure of the curvatures, the shortest distance between nodes can be obtained by considering only 3, 4 or 5-cycles for sparse large graphs (see **Appendix A**). As shown in Fig. 2, the centrum nodes are critical parts of these cycles and our samplers preserved the centrum nodes in subgraphs. Table 2 shows that our samplers maximized the exact curvature.

**Table 2.** Comparison of exact Ollivier-Ricci curvature using real-world graphs in [40-42]. Self-loops, multi-edges and edge-direction of the graphs have been removed.

| Exact Ollivier-Ricci curvature (see Appendix A) | | Average of edge curvatures of 1,000 subgraphs with 10% partial nodes. (The curvature of an edge refers to its curvature on the original graph, not on a subgraph) | | | | | Original Graph |
|---|---|---|---|---|---|---|---|
| | | GraphSAINT -Edge | GraphSAINT -RW | LoCur | HIS-FF | HIS-RW | |
| Graphs in [42] | ego-Facebook 4,039 nodes/88,234 edges | 0.19 | 0.22 | 0.26 | 0.41 | **0.69** | 0.32 |
| | CA-GrQc 5,242 nodes/14,484 edges | 0.04 | 0.01 | 0.05 | 0.07 | **0.12** | 0.06 |
| | Wiki-Vote 7,115 nodes/100,762 edges | -0.01 | 0.05 | 0.10 | **0.19** | **0.19** | -0.11 |
| | CA-HepTh 9,877 nodes/25,973 edges | -0.30 | -0.15 | -0.07 | **-0.06** | -0.10 | -0.33 |
| | CA-HepPh 12,008 nodes/118,489 edges | 0.11 | 0.16 | 0.21 | 0.56 | **0.64** | 0.16 |
| In [40] | CiteSeer 3,327 nodes/4,732 edges | -0.03 | -0.01 | 0.01 | **0.02** | **0.02** | -0.24 |
| In [41] | Pubmed 19,717 nodes/44,338 edges | -0.62 | -0.45 | -0.44 | **-0.43** | -0.44 | -0.61 |

## 5.3 Variance reduction for node feature aggregation

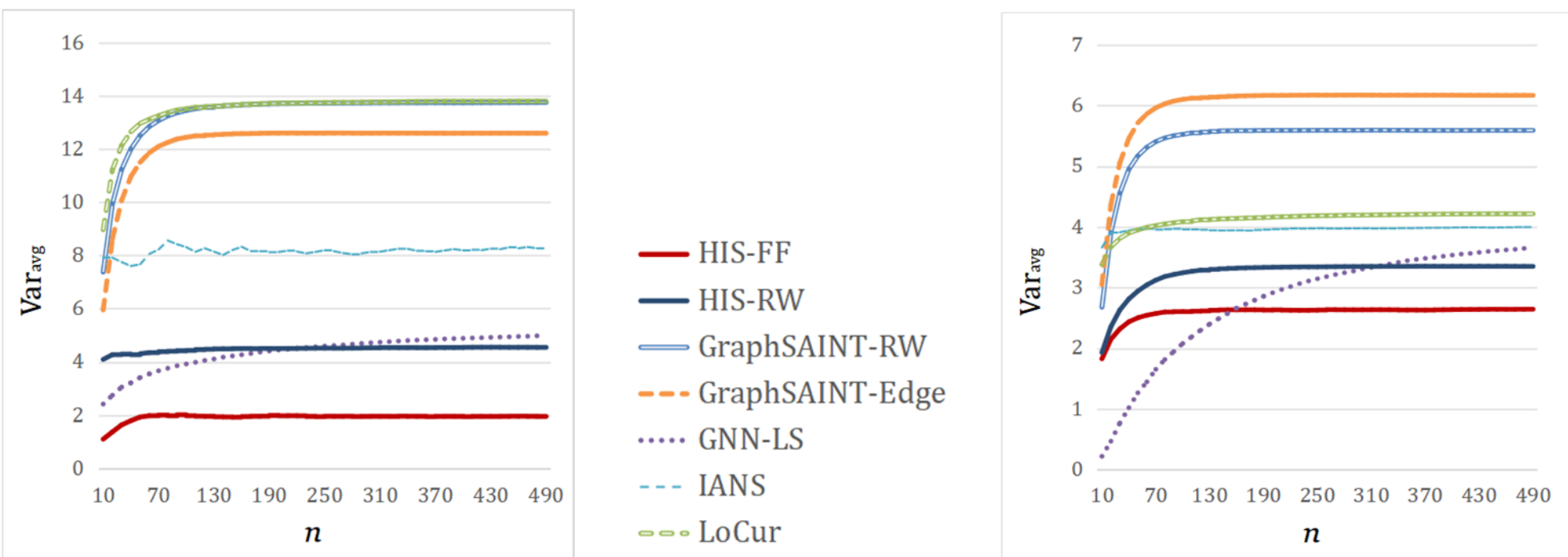

($a$) PPI-Large training graph ($b$) Ogbn-arxiv training graph

**Figure 7.** Comparison of node aggregation variance on the same two-layer GraphSAGE model [4]. Sample size (i.e., the number of nodes in subgraphs) was set as 10% partial nodes of the training graphs. ($a$) $\mathbf{Var}_{\text{avg}}$ *vs.* $n$ in PPI-Large training graph, where $\mathbf{Var}_{\text{avg}}$ is defined in Eq. (26) and $n$ denotes the number of subgraphs. ($b$) $\mathbf{Var}_{\text{avg}}$ *vs.* $n$ in Ogbn-arxiv training graph.

Eqs. (9) to (17) allow a theoretical evaluation of the variance in a single-layer GCN. Moreover, the variance in multi-layer GCNs with non-linear activations can be evaluated experimentally. Let $G_1, G_2, \dots, G_n$ be $n$ subgraphs obtained by a sampler $\mathcal{R}$, and $Y_i(v)$ be 2-norm of the feature vector of node $v$ in subgraph $G_i = (\mathcal{V}_i, \mathcal{E}_i)$ after the forward propagation of a full GCN on $G_i$ with the same initialization weight matrix if $v \in \mathcal{V}_i$.

Subsequently, the feature aggregation variance of node $v$ is calculated as follows:

$$\mathbf{Var}(v) = \frac{1}{k}\sum_{j=1}^{k}\left(Y_{i_j}(v) - \overline{Y}(v)\right)^2 \tag{25}$$

where $Y_{i_1}(v), Y_{i_2}(v), \cdots, Y_{i_k}(v)$ are respectively derived by the forward propagation on $G_{i_1}, G_{i_2}, \cdots$, $G_{i_k}$ that are all subgraphs preserving node $v$, and $\overline{Y}(v) = \frac{1}{k}\sum_{j=1}^{k} Y_{i_j}(v)$.

In addition, the average value of the feature aggregation variance for all the nodes in the training graph $G = (\mathcal{V}, \mathcal{E})$ is calculated as follows:

$$\mathbf{Var}_{\mathrm{avg}} = \frac{\sum_{v\in\mathcal{V}} \mathbf{Var}(v)}{|\mathcal{V}|} \tag{26}$$

An experimental comparison of the node aggregation variance is shown in Fig. 7, which verifies that the empirical probability distribution $q(u' \mid v)$ defined in Eq. (23) does not influence the effectiveness of our method $\mathrm{HIS}_{\mathrm{GCNs}}$ in variance reduction. Low variance accelerates GCN convergence; thus, experiments on GCN convergence are supplemented in Section 5.6.1.

### 5.4 Settings for node classification tasks

The node classification tasks were tested on the six open datasets listed in Table 1. For all datasets, we performed a grid search on the hyperparameter space defined by:

- Dropout: {0.1, 0.15, 0.2, 0.25, 0.3, 0.4, ⋯, 0.8}
- Hidden dimension: {128, 256, 512}
- Learning rate: {0.1, 0.02, 0.01, 0.001, 0.0001}

For our methods HIS-FF in **Algorithm 1** and HIS-RW in **Algorithm 2**, we used sampling rate $\eta$ to determine the input sample size $\hat{n} = |\mathcal{V}| \cdot \eta$ where $\mathcal{V}$ is the node set of the training graph, and performed grid search for $\eta$ and another input parameter $\gamma$ on the hyperparameter space defined by:

- Sampling rate $\eta$: {0.005, 0.01, 0.02, 0.05}
- Parameter $\gamma$: {0.1, 0.2, 0.3, ⋯, 0.8, 0.9, 1.0}

Fig. 8 and Tables 4 – 7 show the convergence and F1-micro score comparisons of the various methods. All results correspond to the two-layer GNN models. For a specific dataset, we maintained the hidden dimension the same across all methods that was adopted by prior studies [6,32]. Based on the experiments in **Appendix G**, the hidden layer dimension was set to 512 for CiteSeer, Pubmed, PPI-Large and Reddit, and 256 for OGBN-Arxiv and OGBN-Products.

For all the methods, the optimizer and activation functions were Adam and ReLU, respectively. We utilized cross-entropy loss for the single-label classification datasets and binary cross-entropy loss for the multi-task datasets. In addition, we sampled 100 subgraphs per epoch for all the datasets, and the number of epochs was adjusted based on convergence.

The hyperparameters of the HIS-FF and HIS-RW are listed in Table 3. In addition, experimental details for core-periphery partition, parameter $\gamma$, and parameter $\eta$ are presented in **Appendix B.1**, **Appendix B.2**, and **Appendix B.3**, respectively. We first compared sampling-based GCN or GNN methods, including node-wise, layer-wise, and subgraph-based samplings, and then focused on the comparison of graph sampling methods within the same GNN architecture. In the former comparison (Section 5.5), all methods used two-layer GNNs and the same hidden layer dimensions mentioned above; however, their other hyperparameters were optimal and obtained by the searching strategies reported in related papers. In the latter comparison (Section 5.6), the same two-layer GNN model was provided for diverse graph sampling methods, and the model was concretized using the GCN [1], GraphSAGE [4] and GAT [16] in sequence. The optimal baseline is discussed in **Appendix D**.

In addition, experiments of the samplers on $k$-layer GNN models with $k \geq 3$ are presented in **Appendix E**. During training on the deep models, we applied DropEdge to mitigate over-smoothing [43], which isolates output representations from input features as the model depth increases.

**Table 3.** Training and sampling hyperparameters for $HIS_{GCNs}$.

| Sampler | Dataset | Training | | Sampling | | |
|---|---|---|---|---|---|---|
| | | Learning rate | Dropout | Sampling rate $\eta$ | Walk length | Parameter $\gamma$ |
| HIS-FF | CiteSeer | 0.001 | 0.8 | 0.01 | — | 1.0 |
| | Pubmed | 0.0001 | 0.2 | 0.005 | — | 1.0 |
| | PPI-Large | 0.02 | 0.1 | 0.05 | — | 1.0 |
| | Ogbn-arxiv | 0.001 | 0.3 | 0.05 | — | 1.0 |
| | Reddit | 0.001 | 0.15 | 0.02 | — | 0.4 |
| | Ogbn-products | 0.001 | 0.3 | 0.02 | — | 0.4 |
| HIS-RW | CiteSeer | 0.001 | 0.8 | 0.01 | 4 | 1.0 |
| | Pubmed | 0.0001 | 0.2 | 0.005 | 3 | 1.0 |
| | PPI-Large | 0.01 | 0.1 | 0.05 | 15 | 1.0 |
| | Ogbn-arxiv | 0.001 | 0.3 | 0.05 | 2 | 1.0 |
| | Reddit | 0.001 | 0.15 | 0.02 | 4 | 0.4 |
| | Ogbn-products | 0.001 | 0.3 | 0.02 | 2 | 0.4 |

## 5.5 Comparison of sampling-based GCN or GNN methods

Our sampling-based methods, HIS-FF and HIS-RW, were compared with state-of-the-art and open-source methods, including GCN [1], GraphSAGE [4], FastGCN [24], ClusterGCN [27], and SHADOW-SAGE [28], where SHADOW-SAGE is SHADOW-GNN on the GraphSAGE model and has two extractors, namely $L$-hop and Personalized PageRank (PPR)-based [28].

**Table 4.** Comparison of test set F1-micro score on two-layer GNNs.

| Methods | CiteSeer | Pubmed | PPI-Large | Ogbn-arxiv | Reddit | Ogbn-products |
|---|---|---|---|---|---|---|
| GCN | $0.711 \pm 0.007$ | $0.786 \pm 0.002$ | $0.482 \pm 0.005$ | $0.702 \pm 0.003$ | $0.952 \pm 0.004$ | $0.757 \pm 0.002$ |
| GraphSAGE | $0.698 \pm 0.006$ | $0.814 \pm 0.001$ | $0.618 \pm 0.004$ | $0.715 \pm 0.003$ | $0.953 \pm 0.001$ | $0.760 \pm 0.003$ |
| FastGCN | $0.704 \pm 0.015$ | $0.863 \pm 0.015$ | $0.507 \pm 0.028$ | $0.682 \pm 0.043$ | $0.924 \pm 0.012$ | $0.744 \pm 0.005$ |
| ClusterGCN | $0.694 \pm 0.007$ | $0.882 \pm 0.003$ | $0.902 \pm 0.002$ | $0.665 \pm 0.001$ | $0.954 \pm 0.002$ | $0.773 \pm 0.001$ |
| SHADOW-SAGE (2-hop) | $0.726 \pm 0.006$ | $0.894 \pm 0.001$ | $0.948 \pm 0.003$ | $0.715 \pm 0.002$ | $0.966 \pm 0.001$ | $0.787 \pm 0.004$ |
| SHADOW-SAGE (PPR) | $0.727 \pm 0.004$ | $0.896 \pm 0.003$ | $0.964 \pm 0.002$ | $0.721 \pm 0.001$ | $\mathbf{0.967 \pm 0.003}$ | $0.785 \pm 0.003$ |
| HIS-FF | $\mathbf{0.740 \pm 0.007}$ | $\mathbf{0.898 \pm 0.002}$ | $\mathbf{0.985 \pm 0.002}$ | $\mathbf{0.723 \pm 0.001}$ | $\mathbf{0.967 \pm 0.001}$ | $\mathbf{0.788 \pm 0.003}$ |
| HIS-RW | $0.728 \pm 0.012$ | $\mathbf{0.898 \pm 0.002}$ | $0.954 \pm 0.005$ | $0.722 \pm 0.002$ | $\mathbf{0.967 \pm 0.002}$ | $0.779 \pm 0.002$ |

Table 4 compares the $HIS_{GCNs}$ (using the GraphSAGE model) with various baselines; the results were obtained by repeating the experiments five times. In contrast to baselines that operate on arbitrary graphs, including mesh grid, Erdös-Rényi random graphs [33], and power-law graphs [7,8,12], our method $HIS_{GCNs}$ depends on the unique core-periphery structure of scale-free graphs. However, many open standard datasets for node classification are scale-free, such as social, biological, citation, and product information networks. Based on the prior topological characteristics of these networks, our HIS-FF and HIS-RW samplers can easily preserve important information propagation paths in small subgraphs, which helps to improve the F1-micro score for node classification.

## 5.6 Comparison of graph sampling methods

Based on **Algorithms 1 and 2**, the proposed HIS-FF and HIS-RW samplers could be extended to diverse GNN models. Thus, three widely used GNN models, GCN [1], GraphSAGE [4] and GAT [16], were used for comparison. Our samplers were compared with state-of-the-art graph samplers, including GraphSAINT-Edge [6], GraphSAINT-RW [6], LoCur [32], GNN-LS [31], IANS [30], and SLSR [44]. The samplers were tested using the same two-layer GNN model.

### 5.6.1 Convergence and training time comparison

Fig. 8 shows the convergence and training time comparison of various graph samplers for the same two-layer GraphSAGE model. The training and sampling hyperparameters of the samplers are presented in Table 3 and **Appendix D**.

The training time in Fig. 8 corresponds to the GPU execution time, which excludes the time required for data loading, preprocessing, sampling, validation set evaluation and model saving. The cost of preprocessing and sampling is detailed in **Appendix B.1** and **Appendix F**, respectively. All samplers were terminated after a certain number of epochs based on convergence.

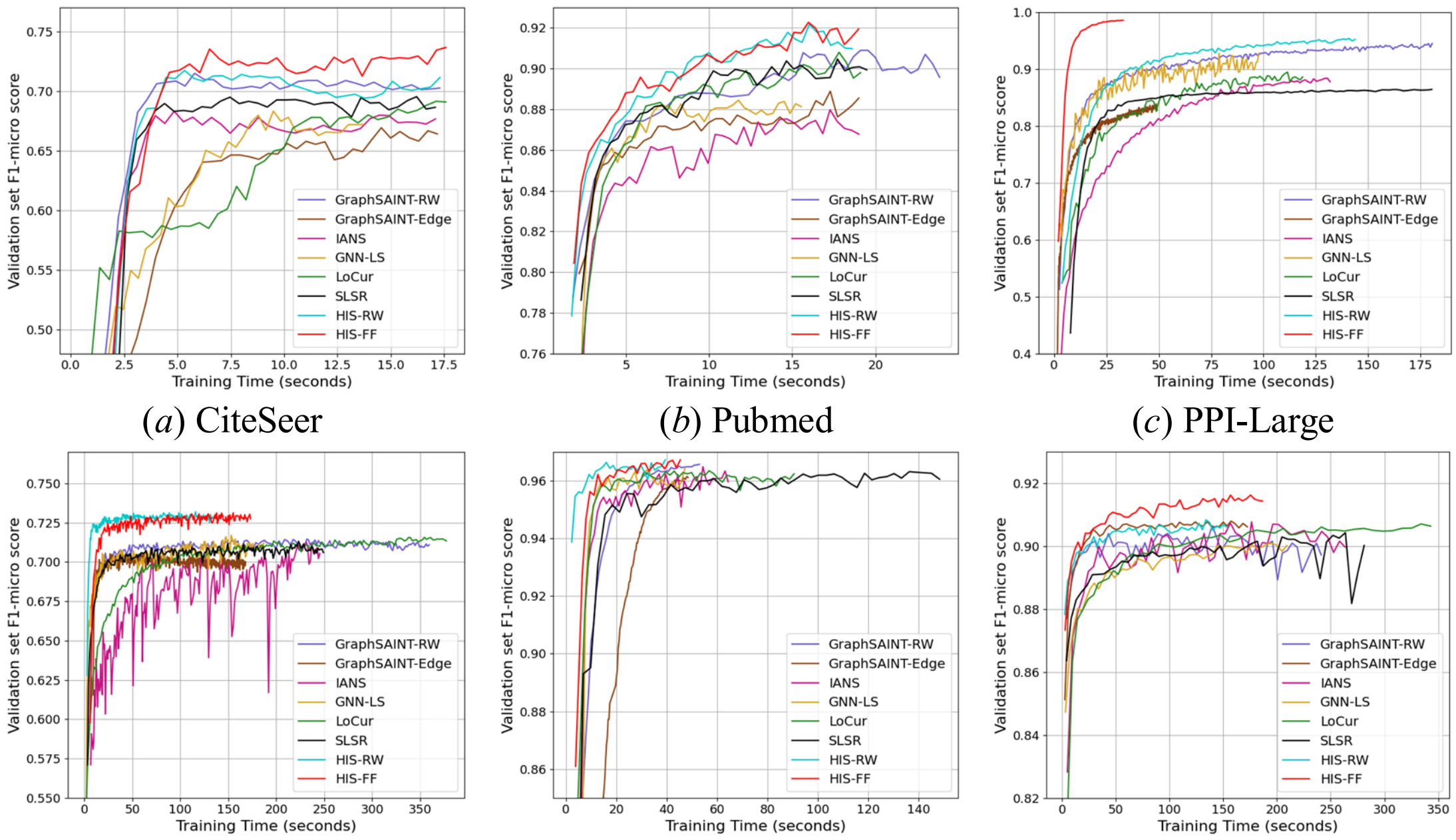

(*a*) CiteSeer (*b*) Pubmed (*c*) PPI-Large

(*d*) OGBN-arxiv (*e*) Reddit (*f*) OGBN-products

**Figure 8.** Convergence curves of samplers on the same two-layer GraphSAGE model.

SLSR [44] is a sampler for network crawling and visualization, aiming to preserve a single subgraph that minimizes variance in the degree property while maintaining the global structures of the original graph, such as the degree distribution, clustering coefficient and path length distribution. Thus, SLSR requires additional time to finely sample only one subgraph, resulting in a loss of diversity in information propagation chains. Deep graph learning propagates information along critical long chains and focuses on local structures, such as edge curvature, because the features in the starting and ending nodes of a path composed of high curvature edges are close. In addition, SLSR does not consider variance in node feature aggregation. Therefore, the difference in goals between network crawling (visualization) and deep graph learning results in SLSR performing poorly in terms of convergence and F1-micro score for node classification tasks, as shown in Fig. 8 and Tables 5 – 7.

Existing subgraph samplers for learning have significantly contributed to the understanding of node aggregation variance and classification accuracy on arbitrary graphs. For scale-free graphs that widely exist in real-world, the role of centrum nodes for deep graph learning is weakened in theoretical analysis: they are key components in forming 5-cycles (or less) that help shorten the length of propagation paths and maximize the curvature. However, their strong interference causes the loss of massive long chains that are composed entirely of low-degree nodes, as shown in Fig. 2. Our samplers, HIS-FF and HIS-RW, maintain the favorable role of the centrum in forming cycles, avoid the unfavorable interference role of the centrum, consider variance reduction for node feature aggregation, reduce the overhead in subgraph sampling through randomness, and enhance the diversity of the long propagation paths in many subgraphs. Although our HIS-FF and HIS-RW samplers did not preserve the degree distribution of the training graph, they were competent for minibatch learning along the propagation paths in scale-free graphs.

Our $HIS_{GCNs}$ method first samples nodes from the sparse periphery, then selects the core nodes based on the previously sampled periphery nodes. The process from the periphery to the core avoids the high uncertainty of randomly extracting the next node from the neighbors of the centrum node of the core. In addition, $HIS_{GCNs}$ preserves the centrum for most subgraphs based on **P3**, which helps control the uncertainty in the subgraph structures. The low node aggregation variance in Fig. 7 provides evidence of the fast convergence of the HIS-FF and HIS-RW samplers, as shown in Fig. 8.

**5.6.2 Test set F1-micro score comparison**

Using the same hyperparameters as in Section 5.6.1, Tables 5 – 7 present the F1-micro score comparison of various samplers on the GCN, GraphSAGE and GAT models. The final F1-scores and their confidence intervals were calculated by repeating the experiments five times using the same device with the same configuration. The maximum number of epochs was set for each dataset. The maximum number was set to 50 for CiteSeer, 100 for Pubmed and Reddit, and 200 for PPI-Large, OGBN-arxiv and OGBN-products. The convergence curves shown in Fig. 8 were determined using the data obtained after running the maximum number of epochs.

**Table 5.** Comparison of test set F1-micro score with different samplers on the same two-layer GCN model. Hyperparameters can be found in Table 3 and **Appendix D**.

| Samplers | CiteSeer | Pubmed | PPI-Large | Ogbn-arxiv | Reddit | Ogbn-products |
|---|---|---|---|---|---|---|
| GraphSAINT -Edge | 0.687 ± 0.007 | 0.866 ± 0.002 | 0.834±0.001 | 0.704±0.002 | 0.962±0.005 | 0.755±0.003 |
| GraphSAINT -RW | 0.726 ± 0.005 | 0.877 ± 0.001 | 0.938±0.005 | 0.711±0.002 | **0.966±0.001** | 0.766±0.003 |
| LoCur | 0.702 ± 0.008 | 0.883 ± 0.003 | 0.890±0.001 | 0.709±0.003 | 0.963±0.002 | 0.755±0.002 |
| GNN-LS | 0.693 ± 0.006 | 0.859± 0.004 | 0.932±0.008 | 0.705±0.006 | 0.965±0.002 | 0.759±0.005 |
| IANS | 0.691 ± 0.015 | 0.860 ± 0.008 | 0.873 ± 0.005 | 0.703±0.003 | 0.962±0.001 | 0.774±0.002 |
| SLSR | 0.699 ± 0.006 | 0.873 ± 0.006 | 0.862±0.003 | 0.702±0.002 | 0.960±0.001 | 0.770±0.004 |
| HIS-FF | **0.743 ± 0.008** | **0.884 ± 0.005** | **0.965±0.004** | **0.719±0.003** | 0.965±0.002 | **0.781±0.001** |
| HIS-RW | 0.735 ± 0.008 | **0.884 ± 0.004** | 0.949±0.003 | 0.716±0.001 | **0.966±0.001** | 0.774±0.002 |

**Table 6.** Comparison of test set F1-micro score with different samplers on the same two-layer GraphSAGE model.

| Samplers | CiteSeer | Pubmed | PPI-Large | Ogbn-arxiv | Reddit | Ogbn-products |
|---|---|---|---|---|---|---|
| GraphSAINT -Edge | 0.690 ± 0.008 | 0.880 ± 0.002 | 0.839±0.005 | 0.705±0.004 | 0.962±0.004 | 0.760±0.005 |
| GraphSAINT -RW | 0.726 ± 0.004 | 0.891 ± 0.002 | 0.940±0.001 | 0.710±0.002 | 0.966±0.003 | 0.773±0.002 |
| LoCur | 0.708 ± 0.008 | 0.897 ± 0.003 | 0.892±0.004 | 0.711±0.002 | 0.964±0.001 | 0.758±0.004 |
| GNN-LS | 0.694 ± 0.007 | 0.873 ± 0.005 | 0.927±0.008 | 0.705±0.005 | 0.964±0.002 | 0.752±0.007 |
| IANS | 0.693 ± 0.018 | 0.878 ± 0.007 | 0.884±0.003 | 0.702±0.002 | 0.962±0.001 | 0.774±0.003 |
| SLSR | 0.701 ± 0.007 | 0.888 ± 0.002 | 0.875±0.004 | 0.705±0.002 | 0.962±0.002 | 0.774±0.004 |
| HIS-FF | **0.740 ± 0.007** | **0.898 ± 0.002** | **0.985±0.002** | **0.723±0.001** | **0.967±0.001** | **0.788±0.003** |
| HIS-RW | 0.728 ± 0.012 | **0.898 ± 0.002** | 0.954±0.005 | 0.722±0.002 | **0.967±0.002** | 0.779±0.002 |

**Table 7.** Comparison of test set F1-micro score with different samplers on the same two-layer GAT model.

| Samplers | CiteSeer | Pubmed | PPI-Large | Ogbn-arxiv | Reddit | Ogbn-products |
|---|---|---|---|---|---|---|
| GraphSAINT -Edge | 0.699 ± 0.005 | 0.867 ± 0.002 | 0.840±0.003 | 0.708±0.002 | 0.963±0.002 | 0.756±0.001 |
| GraphSAINT -RW | 0.736 ± 0.006 | 0.884 ± 0.002 | 0.945±0.002 | 0.714±0.003 | 0.966±0.001 | 0.771±0.002 |
| LoCur | 0.715 ± 0.008 | 0.882 ± 0.002 | 0.902±0.002 | 0.710±0.002 | 0.963±0.001 | 0.759±0.002 |
| GNN-LS | 0.700 ± 0.004 | 0.853 ± 0.004 | 0.937±0.005 | 0.708±0.007 | 0.965±0.001 | 0.758±0.004 |
| IANS | 0.697 ± 0.018 | 0.864 ± 0.008 | 0.858±0.002 | 0.707±0.002 | 0.963±0.001 | 0.776±0.002 |
| SLSR | 0.711 ± 0.006 | 0.874 ± 0.001 | 0.878±0.003 | 0.705±0.002 | 0.963±0.001 | 0.774±0.003 |
| HIS-FF | **0.755 ± 0.005** | 0.885 ± 0.002 | **0.979±0.002** | **0.726±0.002** | **0.967±0.001** | **0.789±0.001** |
| HIS-RW | 0.748 ± 0.011 | **0.892 ± 0.001** | 0.953±0.002 | 0.721±0.002 | **0.967±0.003** | 0.778±0.003 |

According to Tables 5 – 7, the stable F1-micro scores across the different GNN models indicate the high flexibility of all samplers. The datasets in the tables correspond to scale-free networks; thus, using prior information unique to these networks can help improve the F1-micro score.

The power law $P(k) \propto k^{-\tau}$ results in a large number of low-degree periphery nodes and a small number of high-degree core nodes. Additionally, preferential attachment causes the periphery nodes to be densely connected to centrum nodes, which are the core nodes with top highest-degrees. The gap of node degrees between the centrum and marginal core is typically enlarged in large-scale graphs based on preferential attachment. Based on the BA model [7], the edges join a scale-free graph in a chronological order. During the temporal evolution process, a few nodes added to the graph in the early stages are likely to be densely interconnected, forming a prototype of the dense core of a large-scale graph. $HIS_{GCNs}$ adopts a unique structure of scale-free graphs resulting from temporal evolution and preferential attachment. In addition, because two low-degree nodes are likely to influence each other [6], $HIS_{GCNs}$ retains an increasing number of longer chains in the periphery.

Furthermore, as the distance of node features in the starting and ending nodes of a path along high-curvature edges is small [32], $HIS_{GCNs}$ preserves the centrum (which is important for shortening the path length and forming cycles) in most subgraphs. Therefore, the proposed samplers, HIS-FF and HIS-RW, performed better in terms of both F1-micro score and training time for node classification tasks using minibatch learning.

## 6 Conclusions

Existing subgraph samplers have established a superior foundation for minibatch learning on arbitrary graphs, including mesh grids, Erdös-Rényi random graphs, and power-law graphs. Based on this foundation, we only consider minibatch learning on scale-free training graphs with a power-law degree distribution, because these graphs are abundant in real-world and a single research object can utilize its unique structural information.

The advantages of $HIS_{GCNs}$ are summarized as follows: maintaining the favorable role of the centrum in forming 5-cycles (or less), avoiding the unfavorable interference role of the centrum that weakens the preservation of long chains composed entirely of low-degree nodes, and sampling nodes from the periphery to the core to reduce the high uncertainty of randomly extracting the next node from the neighbors of a centrum node. In addition enhances the diversity of propagation paths while minimizing overhead through randomness and considering variance reduction in periphery sampling for node feature aggregation. $HIS_{GCNs}$ maximizes the Ollivier-Ricci curvatures of the edges within a subgraph, preserving critical chains for deep information propagation. It achieves high convergence speed and demonstrates superior performance in both F1-micro score and training time for semi-supervised node classification tasks.

## Acknowledgments

We thank the anonymous reviewers for their comments and suggestions, which helped improve this paper. This study was supported by the Natural Science Foundation of Fujian Province of China (No. 2022J01047).

## References

[1] Kipf, T. N., & Welling, M. (2016). Semi-supervised classification with graph convolutional networks. *arxiv preprint arxiv:1609.02907*.

[2] Neiva, M. B., & Bruno, O. M. (2023). Exploring ordered patterns in the adjacency matrix for improving machine learning on complex networks. *Physica A: Statistical Mechanics and its Applications*, 626, pp. 129086.

[3] Bai, J., Ren, Y., & Zhang, J. (2021). Ripple walk training: A subgraph-based training framework for large and deep graph neural network. In *2021 International Joint Conference on Neural Networks* (IJCNN), IEEE, pp. 1-8.

[4] Hamilton, W., Ying, Z., & Leskovec, J. (2017). Inductive representation learning on large graphs. *Advances in Neural Information Processing Systems*, pp. 1024-1034.

[5] Serafini, M., & Guan, H. (2021). Scalable graph neural network training: The case for sampling. *ACM SIGOPS Operating Systems Review*, 55(1), pp. 68-76.

[6] Zeng, H., Zhou, H., Srivastava, A., Kannan, R., & Prasanna, V. (2020). GraphSAINT: Graph sampling based inductive learning method. In *International conference on learning representations*.

[7] Barabási, A. L., & Albert, R. (1999). Emergence of scaling in random networks. *Science*, 286(5439), pp. 509-512.

[8] Kojaku, S., & Masuda, N. (2018). Core-periphery structure requires something else in the network. *New Journal of physics*, 20(4), pp. 043012.

[9] Kumar, R., Gurugubelli, S., & Chepuri, S. P. (2022). Identifying core-periphery structures using graph neural networks. In *2022 56th Asilomar Conference on Signals, Systems, and Computers*, IEEE, pp. 251-255.

[10] Gurugubelli, S., & Chepuri, S. P. (2022). Generative models and learning algorithms for core-periphery structured graphs. *arxiv preprint arxiv:2210.01489*.

[11] Bansal, N., Kaouri, K., & Woolley, T. E. (2025). Reducing size bias in sampling for infectious disease spread on networks. *arxiv preprint arxiv:2501.13195*.

[12] Jiao, B., Lu, X., Xia, J., Gupta, B. B., Bao, L., & Zhou, Q. (2023). Hierarchical sampling for the visualization of large scale-free graphs. *IEEE Transactions on Visualization and Computer Graphics*, 29(12), pp. 5111-5123.

[13] Xia, F., Sun, K., Yu, S., Aziz, A., Wan, L., Pan, S., & Liu, H. (2021). Graph learning: A survey. *IEEE Transactions on Artificial Intelligence*, 2(2), pp. 109-127.

[14] Yu, J., Rui, Y., & Tao, D. (2014). Click prediction for web image reranking using multimodal sparse coding. *IEEE Transactions on image processing*, 23(5), pp. 2019-2032.

[15] Zhang, Y., Liu, B., Bao, J., Huang, Q., Zhang, M., & Yu, J. (2024). Learnability matters: Active learning for video captioning. *Advances in Neural Information Processing Systems*, 37, pp. 37928-37954.

[16] Veličković, P., Cucurull, G., Casanova, A., Romero, A., Liò, P., & Bengio, Y. (2018). Graph Attention Networks. In *International Conference on Learning Representations*.

[17] Li, D., Wang, Z., Chen, Y., Jiang, R., Ding, W., & Okumura, M. (2025). A survey on deep active learning: Recent advances and new frontiers. *IEEE Transactions on Neural Networks and Learning Systems*, 36(4), pp. 5879-5899.

[18] Ding, Y., Zhang, Z., Zhao, X., Hong, D., Cai, W., Yu, C., Yang, N., & Cai, W. (2022). Multi-feature fusion: Graph neural network and CNN combining for hyperspectral image classification. *Neurocomputing*, 501, pp. 246-257.

[19] Su, P., Xiang, C., & Chen, D. (2025). Adopting graph neural networks to understand and reason about dynamic driving scenarios. *IEEE Open Journal of Intelligent Transportation Systems*, 6, pp. 579-589.

[20] Das, S. S., Ferdous, S. M., Halappanavar, M. M., Serra, E., & Pothen, A. (2024). Ags-gnn: Attribute-guided sampling for graph neural networks. In *Proceedings of the 30th ACM SIGKDD Conference on Knowledge Discovery and Data Mining*, pp. 538-549.

[21] Younesian, T., Daza, D., van Krieken, E., Thanapalasingam, T., & Bloem, P. (2023). Grapes: Learning to sample graphs for scalable graph neural networks. *arxiv preprint arxiv:2310.03399*.

[22] Ying, R., He, R., Chen, K., Eksombatchai, P., Hamilton, W. L., & Leskovec, J. (2018). Graph convolutional neural networks for web-scale recommender systems. In *Proceedings of the 24th ACM SIGKDD international conference on knowledge discovery & data mining*, pp. 974-983.

[23] Chen, J., Zhu, J., & Song, L. (2018). Stochastic training of graph convolutional networks with variance reduction. In *International Conference on Machine Learning*, PMLR, pp. 942-950.

[24] Chen, J., Ma, T., & Xiao, C. (2018). FastGCN: Fast learning with graph convolutional networks via importance sampling. In *International Conference on Learning Representations*, ICLR.

[25] Huang, W., Zhang, T., Rong, Y., & Huang, J. (2018). Adaptive sampling towards fast graph representation learning. *Advances in Neural Information Processing Systems*, pp. 4558-4567.

[26] Zou, D., Hu, Z., Wang, Y., Jiang, S., Sun, Y., & Gu, Q. (2019). Layer-dependent importance

sampling for training deep and large graph convolutional networks. *Advances in Neural Information Processing Systems*, pp. 11249-11259.

[27] Chiang, W. L., Liu, X., Si, S., Li, Y., Bengio, S., & Hsieh, C. J. (2019). Cluster-gcn: An efficient algorithm for training deep and large graph convolutional networks. In *Proceedings of the 25th ACM SIGKDD international conference on knowledge discovery & data mining*, pp. 257-266.

[28] Zeng, H., Zhang, M., Xia, Y., Srivastava, A., Malevich, A., Kannan, R., Prasanna, V., Jin, L., & Chen, R. (2021). Decoupling the depth and scope of graph neural networks. *Advances in Neural Information Processing Systems*, pp. 19665-19679.

[29] Wan, C., Li, Y., Li, A., Kim, N. S., & Lin, Y. (2022). BNS-GCN: Efficient full-graph training of graph convolutional networks with partition-parallelism and random boundary node sampling. In *Proceedings of Machine Learning and Systems*, pp. 673-693.

[30] Zhang, Q., Sun, Y., Hu, Y., Wang, S., & Yin, B. (2023). A subgraph sampling method for training large-scale graph convolutional network. *Information Sciences*, 649, pp. 119661.

[31] Zhao, W., Guo, T., Yu, X., & Han, C. (2023). A learnable sampling method for scalable graph neural networks. *Neural Networks*, 162, pp. 412-424.

[32] Shu, D. W., Kim, Y., & Kwon, J. (2023). Localized curvature-based combinatorial subgraph sampling for large-scale graphs. *Pattern Recognition*, 139, pp. 109475.

[33] Erdös, P., & Rényi, A. (1960). On the evolution of random graphs. *Publ. Math. Inst. Hungar. Acad. Sci*, 5, pp. 17-61.

[34] Qi, X. (2024). Weighted jump in random walk graph sampling. *Neurocomputing*, 586, pp. 127581.

[35] Leskovec, J., Kleinberg, J., & Faloutsos, C. (2005). Graphs over time: densification laws, shrinking diameters and possible explanations. In *Proceedings of the eleventh ACM SIGKDD international conference on Knowledge discovery in data mining*, pp. 177-187.

[36] Yin, H., Shao, Y., Miao, X., Li, Y., & Cui, B. (2022). Scalable graph sampling on gpus with compressed graph. In *Proceedings of the 31st ACM International Conference on Information & Knowledge Management*, pp. 2383-2392.

[37] Hu, W., Fey, M., Zitnik, M., Dong, Y., Ren, H., Liu, B., Catasta, M., & Leskovec, J. (2020). Open graph benchmark: Datasets for machine learning on graphs. *Advances in Neural Information Processing Systems*, pp. 22118-22133.

[38] Ollivier, Y. (2009). Ricci curvature of Markov chains on metric spaces. *Journal of Functional Analysis*, 256(3), pp. 810-864.

[39] Ye, Z., Liu, K. S., Ma, T., Gao, J., & Chen, C. (2019). Curvature graph network. In *International conference on learning representations*.

[40] Giles, C. L., Bollacker, K. D., & Lawrence, S. (1998). CiteSeer: An automatic citation indexing system. In *Proceedings of the third ACM conference on Digital libraries*, pp. 89-98.

[41] Yang, Z., Cohen, W., & Salakhudinov, R. (2016). Revisiting semi-supervised learning with graph embeddings. In *International conference on machine learning* (pp. 40-48). PMLR.

[42] Leskovec, J. (2014). SNAP Datasets: Stanford large network dataset collection. Retrieved February 2025 from *http://snap.stanford.edu/data*.

[43] Rong, Y., Huang, W., Xu, T., & Huang, J. (2020). Dropedge: Towards deep graph convolutional networks on node classification. In *International Conference on Learning Representations*.

[44] Jiao, B. (2024). Sampling unknown large networks restricted by low sampling rates. *Scientific Reports*, 14(1), pp. 13340.

## **Appendix A:** Calculation of the exact Ollivier-Ricci curvature [32,38,39]

Define a random walk $\mu$ on a graph $G = (\mathcal{V}, \mathcal{E})$ as a family of probability measure $\mu_v(\cdot)$ on the node set $\mathcal{V}$ for all $v \in \mathcal{V}$. For a node $p \in \mathcal{V}$, the uniform 1-step random walk $\mu$ is expressed by:

$$\mu_v(p) = \begin{cases} \alpha & \text{if } p = v \\ (1-\alpha)/d_v & \text{if } (p, v) \in \mathcal{E} \\ 0 & \text{otherwise} \end{cases} \tag{27}$$

where $\mathcal{E}$ is the edge set of $G$, and $d_v$ denotes the degree of node $v$. We set parameter $\alpha = 0$ in our experiments. Let $d(u, v)$ be the shortest path distance between nodes $u$ and $v$, then the Ollivier-Ricci curvature $\kappa(u, v)$ is defined as follows [39]:

$$\kappa(u, v) = 1 - \frac{W_1(\mu_u, \mu_v)}{d(u, v)} \tag{28}$$

where $W_1(\mu_u, \mu_v)$ denotes the 1-Wasserstein transport distance that is the optimal value of the objective function in the linear optimization problem ($\alpha = 0$):

$$\begin{aligned} \text{minimize} \quad & \sum_{p \in \mathcal{N}_u} \sum_{q \in \mathcal{N}_v} d(p, q)\pi(p, q) \\ \text{subject to} \quad & \sum\nolimits_{p \in \mathcal{N}_u} \pi(p, q) = \frac{1}{d_v}, \text{for } \forall q \in \mathcal{N}_v \\ & \sum\nolimits_{q \in \mathcal{N}_v} \pi(p, q) = \frac{1}{d_u}, \text{for } \forall p \in \mathcal{N}_u \end{aligned} \tag{29}$$

where $\mathcal{N}_u = \{p \in \mathcal{V} | (p, u) \in \mathcal{E}\}$ and $\mathcal{N}_v = \{q \in \mathcal{V} | (q, v) \in \mathcal{E}\}$.

Such a definition captures the behavior that $\kappa(u, v) = 0$ if the random walkers tend to stay at equal distance, $\kappa(u, v) < 0$ if they tend to diverge, and $\kappa(u, v) > 0$ if they tend to converge.

In 1-hop local structure $\{\mathcal{N}_u \cup u\} \cup \{\mathcal{N}_v \cup v\}$ for an edge $(u, v) \in \mathcal{E}$, the distance transported from the neighboring node of $u$ to the neighboring node of $v$ through a 6-cycle is the same as the distance transported through $u$ and $v$. Thus, the optimal transport distance for the local structure can be obtained by considering only 5-cycles or cycles with less edges [32].

## **Appendix B. 1:** Experiments on core-periphery partition [12]

We adopt the core-periphery partition algorithm in [12] because the algorithm only uses a simple graph property, namely node degree, which incurs slight overhead in the training time. In addition, the algorithm maximizes the number of edges connecting a periphery node and a core node, which helps minimize the core node ratio (i.e., $|\mathcal{V}_{cor}|/|\mathcal{V}|$), defined as the ratio of the number of core nodes to the total number of nodes in large-scale graphs with a power-law degree distribution. Minimization of the core node ratio can enhance the diversity of the peripheral chains in multiple subgraphs and assists in preserving the centrum nodes in most subgraphs.

Our method $\text{HIS}_{\text{GCNs}}$ only needs to execute the partition algorithm once during preprocessing. Table 8 lists the degree threshold, core node ratio, and $|\mathcal{E}_{ver}|/|\mathcal{E}|$ obtained by partitioning on the training graphs $G = (\mathcal{V}, \mathcal{E})$, where $\mathcal{E}_{ver}$ is defined in Eq. (7) and the number of edges in $\mathcal{E}_{ver}$ can be maximized using Eq. (4). Table 8 shows the average CPU execution time for the five realizations of the core-periphery partition. Using the additional scale-free graphs extracted from [42], Table 9 verifies the efficiency of the partitioning algorithm.

In [12], the degree threshold $d_{th}$ starts from 1 and then increases in step length 1 until Eq. (4) is satisfied. The test datasets and optimized code of the core-periphery partition algorithm are available at *https://github.com/HuQiaCHN/HIS-GCN.*

**Table 8.** Core-periphery partition on scale-free training graphs, where $d_{th}$ is the degree threshold, $|\mathcal{V}_{cor}|/|\mathcal{V}|$ is the ratio of the number of core nodes to the total number of nodes, and $|\mathcal{E}_{ver}|/|\mathcal{E}|$ is the ratio of the number of edges in $\mathcal{E}_{ver}$ to the total number of edges.

| Training graphs | Nodes | Edges | $d_{th}$ | $\|\mathcal{V}_{cor}\|/\|\mathcal{V}\|$ | $\|\mathcal{E}_{ver}\|/\|\mathcal{E}\|$ | Average CPU execution time (s) |
|---|---|---|---|---|---|---|
| CiteSeer | 1,812 | 1,351 | 2 | 15.45% | 37.23% | 0.00 |
| Pubmed | 18,217 | 37,900 | 8 | 13.61% | 58.90% | 0.01 |
| PPI-Large | 44,906 | 633,198 | 57 | 12.34% | 47.91% | 0.03 |
| Ogbn-arxiv | 90,941 | 369,033 | 23 | 5.68% | 41.50% | 0.04 |
| Reddit | 153,932 | 5,376,619 | 109 | 16.13% | 39.10% | 0.59 |
| Ogbn-products | 196,615 | 5,451,633 | 79 | 17.72% | 47.83% | 0.42 |

**Table 9.** Core-periphery partition on scale-free graphs in [42].
Self-loops, multi-edges and edge-direction of the graphs have been removed.

| Large-scale graphs in [42] | Nodes | Edges | $d_{th}$ | $\|\mathcal{V}_{cor}\|/\|\mathcal{V}\|$ | $\|\mathcal{E}_{ver}\|/\|\mathcal{E}\|$ | Average CPU execution time (s) |
|---|---|---|---|---|---|---|
| web-Google | 875,713 | 4,322,051 | 20 | 8.60% | 55.54% | 0.92 |
| com-youtube | 1,134,879 | 2,987,595 | 39 | 1.60% | 50.92% | 2.13 |
| as-skitter | 1,696,415 | 11,095,298 | 50 | 2.90% | 61.86% | 2.28 |
| wiki-topcats | 1,791,489 | 25,444,207 | 76 | 5.31% | 61.15% | 10.51 |
| wiki-Talk | 2,394,385 | 4,659,565 | 79 | 0.34% | 83.44% | 1.71 |
| cit-Patents | 3,774,768 | 16,518,947 | 13 | 20.06% | 41.35% | 0.32 |

## **Appendix B. 2:** Experiments on parameter $\gamma$

Using the same configuration and hyperparameters (except for parameter $\gamma$) as those in Table 6, we test the relationship between parameter $\gamma$ and test set F1-micro score of HIS-FF on two-layer GraphSAGE model. As shown in Table 10, we experimentally set $\gamma = 0.4$ for training graphs containing over $100{,}000$ nodes, otherwise $\gamma = 1$.

**Table 10.** Relationship between parameter $\gamma$ and test set F1-micro score of HIS-FF on two-layer GraphSAGE model.

| **Citeseer** training graph with 1,812 nodes | | **Pubmed** training graph with 18,217 nodes | | **PPI-Large** training graph with 44,906 nodes | | **Ogbn-arxiv** training graph with 90,941 nodes | | **Reddit** training graph with 153,932 nodes | | **Ogbn-products** training graph with 196,615 nodes | |
|---|---|---|---|---|---|---|---|---|---|---|---|
| $\gamma$ | Test F1-score | $\gamma$ | Test F1-score | $\gamma$ | Test F1-score | $\gamma$ | Test F1-score | $\gamma$ | Test F1-score | $\gamma$ | Test F1-score |
| 0.2 | 0.737±0.002 | 0.2 | 0.897±0.003 | 0.2 | 0.825±0.001 | 0.2 | 0.715±0.003 | 0.2 | 0.965±0.001 | 0.2 | 0.784±0.003 |
| 0.4 | 0.735±0.003 | 0.4 | 0.896±0.005 | 0.4 | 0.954±0.003 | 0.4 | 0.718±0.004 | 0.4 | **0.967±0.001** | 0.4 | **0.788±0.003** |
| 0.6 | 0.739±0.002 | 0.6 | 0.895±0.004 | 0.6 | 0.970±0.002 | 0.6 | 0.719±0.001 | 0.6 | 0.966±0.002 | 0.6 | 0.784±0.001 |
| 0.8 | 0.734±0.003 | 0.8 | 0.895±0.003 | 0.8 | 0.977±0.003 | 0.8 | 0.722±0.002 | 0.8 | 0.963±0.002 | 0.8 | 0.781±0.001 |
| 1 | **0.740±0.007** | 1 | **0.898±0.002** | 1 | **0.985±0.002** | 1 | **0.723±0.001** | 1 | 0.962±0.001 | 1 | 0.777±0.002 |

## **Appendix B. 3:** Experiments on sampling rate $\eta$

Using the same configuration and hyperparameters (except for sampling rate $\eta$) as those in Table 6, we test the relationship among sampling rate $\eta$, test set F1-micro score, and one epoch training time of HIS-FF on the same two-layer GraphSAGE model. As shown in Fig. 9, with increasing $\eta$, the test F1-score slightly decreases after reaching its peak, while the one epoch training time continues to increase. We observed that the test set F1-score peak corresponded to low sampling rates; thus, our method $\text{HIS}_{\text{GCNs}}$ is capable of minibatch learning with a small size.

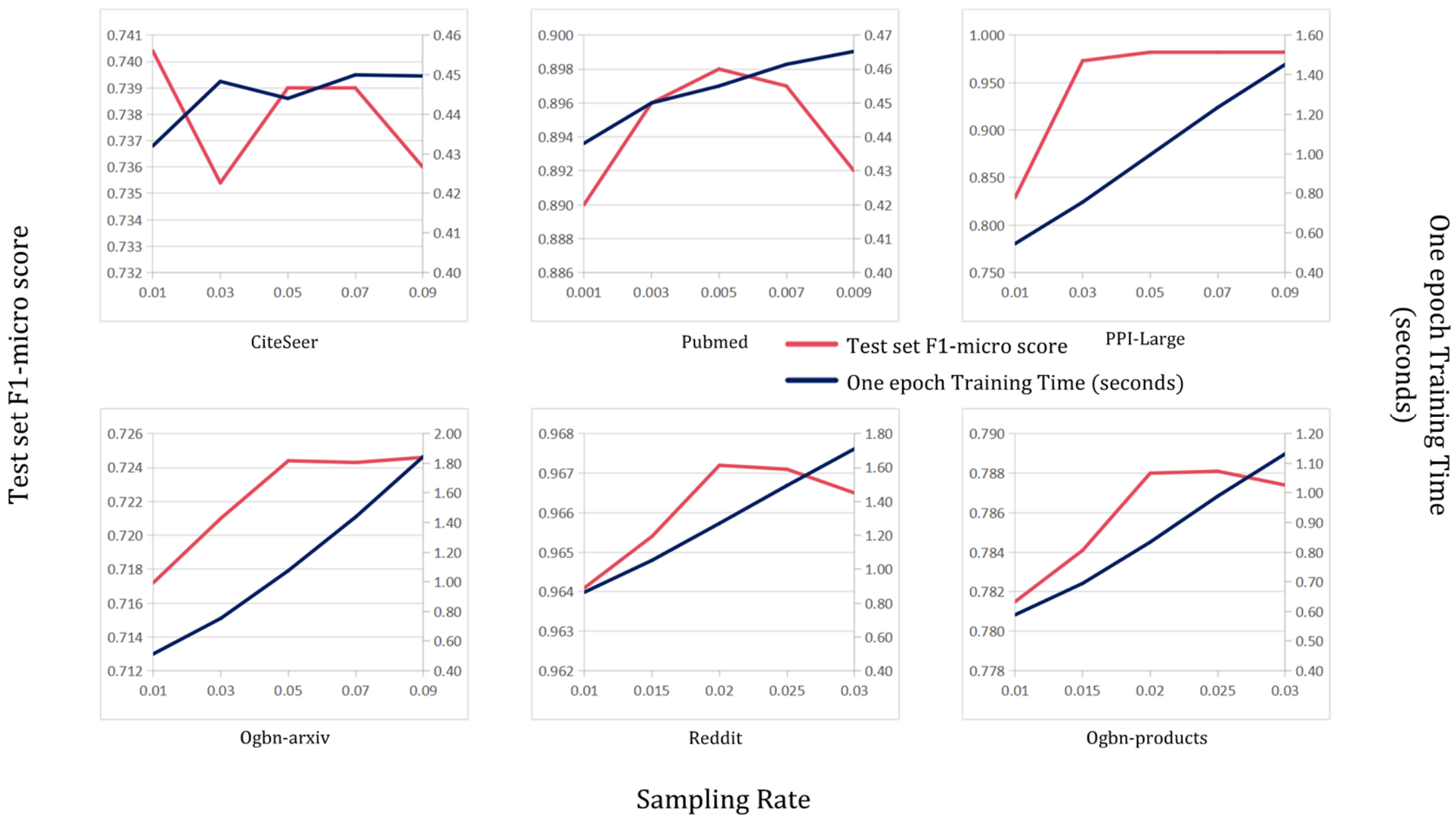

**Figure 9.** Relationship among sampling rate $\eta$, test set F1-micro score, and one epoch training time (seconds) of HIS-FF on the same two-layer GraphSAGE model.
The one epoch training time corresponds to the average time of 100 epochs.

## **Appendix C:** $\text{HIS}_{\text{GCNs}}$ training algorithm

**Algorithm 3:** $\text{HIS}_{\text{GCNs}}$ training algorithm

**Input:** Scale-free training graph $G = (\mathcal{V}, \mathcal{E}, X)$; Labels $\bar{Y}$; Sampler SAMPLE (HIS-FF or HIS-RW).
**Output:** GCN model with trained weights.

1: Pre-processing: Setup SAMPLE parameters; Core-periphery partition [12]; Compute normalization coefficient $\lambda$ [6].
2: **For** each minibatch **do**
3: $G_{sub} = (\mathcal{V}_{sub}, \mathcal{E}_{sub}) \leftarrow$ Sampled subgraph of $G$ according to SAMPLE.
4: GCN construction on $G_{sub}$.
5: $\{y_v | v \in \mathcal{V}_{sub}\} \leftarrow$ Forward propagation of $\{x_v | v \in \mathcal{V}_{sub}\}$.
6: Backward propagation from $\lambda$-normalized loss $L(y_v, \bar{y}_v)$ [6]. Update weights.
7: **End for**

Algorithm 3 illustrates the training algorithm of our method $\text{HIS}_{\text{GCNs}}$.

## **Appendix D:** Training and sampling hyperparameters for baselines

Training and sampling hyperparameters of the baseline data are listed in Table 11. The number of subgraphs used for training was determined by the number of epochs illustrated in Section 5.6.2; thus, they are not listed in Table 11.

**Table 11.** Training and sampling hyperparameters for baselines.
The number of nodes in subgraphs can be determined by sampling rate $\eta$.

| Sampler | Dataset | Training | | Sampling | | | | | | |
|---|---|---|---|---|---|---|---|---|---|---|
| | | Learning rate | Dropout | Sampling rate $\eta$ | Edge budget | Roots | Walk length | Steps | Cluster number | Neighbor sampling ratio |
| GraphSAINT -Edge | CiteSeer | 0.001 | 0.8 | — | 10 | — | — | — | — | — |
| | Pubmed | 0.0001 | 0.2 | — | 80 | — | — | — | — | — |
| | PPI-Large | 0.01 | 0.1 | — | 1600 | — | — | — | — | — |
| | Ogbn-arxiv | 0.001 | 0.3 | — | 2600 | — | — | — | — | — |
| | Reddit | 0.001 | 0.1 | — | 6000 | — | — | — | — | — |
| | Ogbn-products | 0.001 | 0.3 | — | 2800 | — | — | — | — | — |
| GraphSAINT -RW | CiteSeer | 0.001 | 0.8 | — | — | 5 | 4 | — | — | — |
| | Pubmed | 0.0001 | 0.2 | — | — | 25 | 3 | — | — | — |
| | PPI-Large | 0.01 | 0.1 | — | — | 1000 | 4 | — | — | — |
| | Ogbn-arxiv | 0.001 | 0.3 | — | — | 1250 | 2 | — | — | — |
| | Reddit | 0.001 | 0.1 | — | — | 2000 | 4 | — | — | — |
| | Ogbn-products | 0.001 | 0.3 | — | — | 1250 | 2 | — | — | — |
| LoCur | CiteSeer | 0.001 | 0.8 | 0.01 | — | — | — | 4 | — | — |
| | Pubmed | 0.0001 | 0.2 | 0.005 | — | — | — | 3 | — | — |
| | PPI-Large | 0.02 | 0.1 | 0.05 | — | — | — | 4 | — | — |
| | Ogbn-arxiv | 0.001 | 0.3 | 0.05 | — | — | — | 3 | — | — |
| | Reddit | 0.001 | 0.1 | 0.02 | — | — | — | 3 | — | — |
| | Ogbn-products | 0.001 | 0.3 | 0.02 | — | — | — | 3 | — | — |
| GNN-LS | CiteSeer | 0.001 | 0.8 | 0.01 | — | — | — | — | — | — |
| | Pubmed | 0.0001 | 0.2 | 0.005 | — | — | — | — | — | — |
| | PPI-Large | 0.02 | 0.1 | 0.05 | — | — | — | — | — | — |
| | Ogbn-arxiv | 0.001 | 0.3 | 0.05 | — | — | — | — | — | — |
| | Reddit | 0.001 | 0.1 | 0.02 | — | — | — | — | — | — |
| | Ogbn-products | 0.001 | 0.3 | 0.02 | — | — | — | — | — | — |
| IANS | CiteSeer | 0.001 | 0.8 | 0.01 | — | — | — | — | 5 | 0.5 |
| | Pubmed | 0.0001 | 0.2 | 0.005 | — | — | — | — | 10 | 0.5 |
| | PPI-Large | 0.01 | 0.3 | 0.05 | — | — | — | — | 15 | 0.5 |
| | Ogbn-arxiv | 0.001 | 0.3 | 0.05 | — | — | — | — | 10 | 0.5 |
| | Reddit | 0.001 | 0.1 | 0.02 | — | — | — | — | 10 | 0.5 |
| | Ogbn-products | 0.001 | 0.3 | 0.02 | — | — | — | — | 45 | 0.5 |
| SLSR | CiteSeer | 0.001 | 0.8 | 0.01 | — | — | — | — | — | — |
| | Pubmed | 0.0001 | 0.2 | 0.005 | — | — | — | — | — | — |
| | PPI-Large | 0.02 | 0.1 | 0.05 | — | — | — | — | — | — |
| | Ogbn-arxiv | 0.001 | 0.3 | 0.05 | — | — | — | — | — | — |
| | Reddit | 0.001 | 0.1 | 0.02 | — | — | — | — | — | — |
| | Ogbn-products | 0.001 | 0.3 | 0.02 | — | — | — | — | — | — |

## **Appendix E:** Experiments on $k$-layer GNN models with $k \geq 3$

We used the GraphSAGE model for the experiments, as shown in Table 12 and Fig. 10. During training on the $k$-layer models with $k \geq 3$, we apply DropEdge to both the baselines and $\text{HIS}_{\text{GCNs}}$ models. DropEdge [43] helps improve the accuracy by mitigating over-smoothing.

**Table 12.** Comparison of test set F1-micro score on the same 3-layer GraphSAGE model (tuned with DropEdge).

| Samplers | CiteSeer | Pubmed | PPI-Large | Ogbn-arxiv | Reddit | Ogbn-products |
|---|---|---|---|---|---|---|
| GraphSAINT -Edge | $0.724 \pm 0.006$ | $0.898 \pm 0.002$ | $0.834 \pm 0.001$ | $0.705 \pm 0.002$ | $0.964 \pm 0.005$ | $0.766 \pm 0.002$ |
| GraphSAINT -RW | $0.731 \pm 0.003$ | $0.899 \pm 0.004$ | $0.943 \pm 0.002$ | $0.714 \pm 0.002$ | $0.967 \pm 0.001$ | $0.789 \pm 0.005$ |
| LoCur | $0.721 \pm 0.008$ | $0.894 \pm 0.003$ | $0.904 \pm 0.004$ | $0.713 \pm 0.004$ | $0.964 \pm 0.003$ | $0.773 \pm 0.003$ |
| GNN-LS | $0.730 \pm 0.004$ | $0.882 \pm 0.004$ | $0.934 \pm 0.008$ | $0.707 \pm 0.005$ | $0.965 \pm 0.001$ | $0.758 \pm 0.004$ |
| IANS | $0.710 \pm 0.015$ | $0.873 \pm 0.005$ | $0.925 \pm 0.005$ | $0.705 \pm 0.004$ | $0.963 \pm 0.001$ | $0.785 \pm 0.002$ |
| HIS-FF | $\mathbf{0.747 \pm 0.009}$ | $\mathbf{0.902 \pm 0.005}$ | $\mathbf{0.989 \pm 0.004}$ | $\mathbf{0.732 \pm 0.003}$ | $0.967 \pm 0.002$ | $\mathbf{0.794 \pm 0.003}$ |
| HIS-RW | $0.740 \pm 0.006$ | $0.896 \pm 0.004$ | $0.974 \pm 0.003$ | $0.729 \pm 0.001$ | $\mathbf{0.968 \pm 0.001}$ | $0.792 \pm 0.002$ |

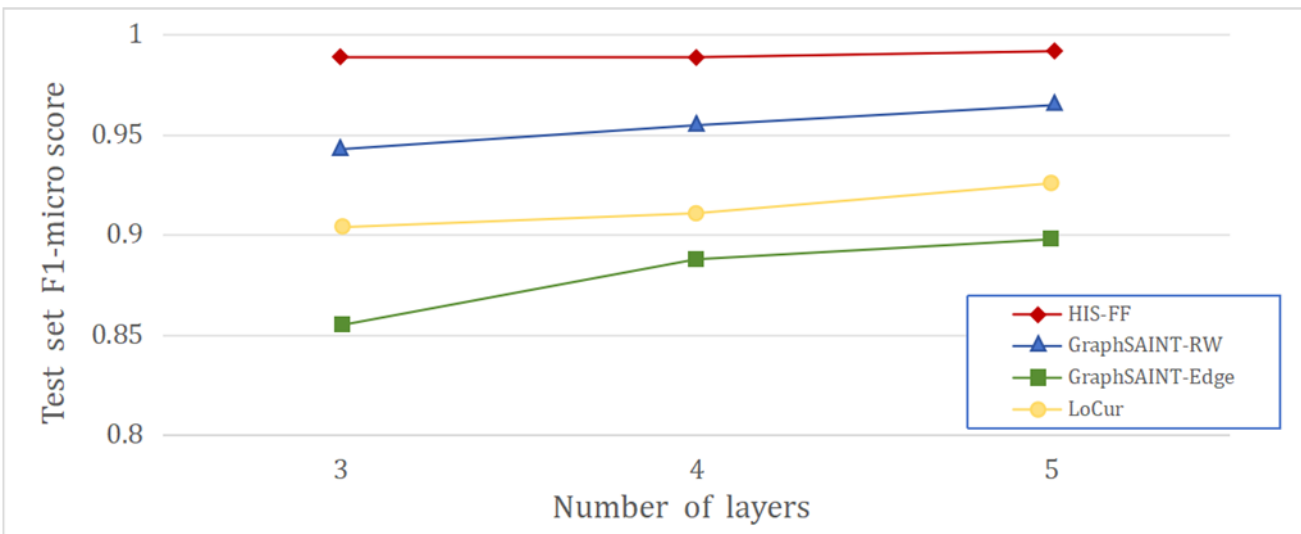

**Figure 10.** Comparison of test set F1-micro score on PPI-Large as model depth increases (tuned with DropEdge), with hidden layer dimension 512 and parameters in Tables 3 and 11.

## **Appendix F:** Cost of graph samplers

Graph sampling introduces a small overhead in training time. Let $t_s$ be the CPU time to sample 100 subgraphs, and $t_t$ be the GPU time to perform forward and backward propagation on one epoch (composed of 100 subgraphs) on the two-layer GraphSAGE model. Table 13 lists the CPU and GPU times for various datasets. None of the samplers employed a parallel strategy.

**Table 13.** Comparison of the CPU time $t_s$ (seconds) to sample 100 subgraphs and the GPU time $t_t$ (seconds) to perform the propagation on one epoch composed of 100 subgraphs.

| Samplers | CiteSeer | | Pubmed | | PPI-Large | | Ogbn-arxiv | | Reddit | | Ogbn-products | |
|---|---|---|---|---|---|---|---|---|---|---|---|---|
| | $t_s$ | $t_t$ | $t_s$ | $t_t$ | $t_s$ | $t_t$ | $t_s$ | $t_t$ | $t_s$ | $t_t$ | $t_s$ | $t_t$ |
| GraphSAINT -Edge | 0.015 | 0.486 | 0.026 | 0.432 | 0.154 | 0.538 | 0.868 | 1.144 | 0.923 | 1.341 | 0.945 | 0.720 |
| GraphSAINT -RW | 0.015 | 0.455 | 0.032 | 0.414 | 0.653 | 0.772 | 1.202 | 0.985 | 1.054 | 1.412 | 1.311 | 0.726 |
| LoCur | 0.044 | 0.764 | 0.191 | 0.678 | 1.484 | 0.969 | 3.526 | 1.410 | 3.345 | 1.377 | 3.682 | 1.271 |
| GNN-LS | 0.025 | 0.460 | 0.058 | 0.344 | 0.225 | 0.489 | 0.929 | 0.893 | 1.011 | 0.593 | 1.233 | 0.591 |
| IANS | 0.015 | 0.450 | 0.045 | 0.446 | 1.224 | 1.207 | 1.501 | 0.905 | 2.320 | 1.310 | 2.623 | 1.069 |
| SLSR | 0.076 | 0.442 | 0.623 | 0.438 | 3.651 | 1.012 | 4.622 | 1.102 | 4.088 | 1.018 | 4.824 | 0.854 |
| HIS-FF | 0.018 | 0.427 | 0.151 | 0.449 | 0.896 | 0.995 | 1.195 | 1.076 | 1.022 | 1.271 | 1.326 | 0.834 |
| HIS-RW | 0.014 | 0.425 | 0.022 | 0.441 | 0.165 | 0.998 | 0.521 | 1.164 | 0.415 | 1.144 | 0.538 | 0.826 |

HIS-FF and HIS-RW are traversal-based samplers that iteratively select the next node from the nodes adjacent to the node being accessed. Thus, the sampling time $t_s$ was proportional to the following factors: number of sampled nodes (factor 1), frequency with which nodes were accessed repeatedly (factor 2), number of nodes adjacent to the node being accessed (factor 3), and additional execution time, excluding the time spent randomly selecting the next node (factor 4). The greater the number of nodes in factor 3, the more time it takes to randomly select the next node from the nodes adjacent to the node being accessed. HIS-FF and HIS-RW traverse on the periphery, and the number of nodes adjacent to each periphery node is limited to below the degree threshold $d_{th}$, which helps to compress the number of nodes in factor 3. In a scale-free graph, the small and dense core can easily be accessed repeatedly. HIS-FF and HIS-RW avoid traversing the core nodes. Thus, our samplers can reduce the impact of factor 2 on computational time. All samplers use low sampling rates corresponding to a small number of sampled nodes in factor 1, which can reduce the computational time.

In contrast to HIS-RW, HIS-FF requires additional time to process the FIFO queue $Q$, generate geometric distribution random variables, and determine whether the adjacent nodes of a node $v$ fall into $\mathcal{N}_{per}^{1}(v) = \{u \in \mathcal{V}_{per} | (u, v) \in \mathcal{E}_{per}, u \notin \mathcal{V}_{sub}, u \notin Q\}$ that consists of nodes that have not been traversed. Because all the samplers in Table 13 are highly efficient, these additional operations associated with factor 4 result in HIS-FF having a time consumption $t_s$ close to that of GraphSAINT-RW. Compared with GraphSAINT-RW, HIS-RW reduced $t_s$ by more than $55\%$ when sampling 100 subgraphs on the large-scale training graphs of PPI-Large, Ogbn-arxiv, Reddit and Ogbn-products.

Based on Fig. 7, HIS-FF exhibits lower node aggregation variance than HIS-RW, as HIS-FF and HIS-RW randomly select the next node from $\mathcal{N}_{per}^{1}(v)$ and $\mathcal{N}_{per}^{2}(v) = \{u \in \mathcal{V}_{per} | (u, v) \in \mathcal{E}_{per}\}$, respectively, and $|\mathcal{N}_{per}^{1}(v)| < |\mathcal{N}_{per}^{2}(v)|$. In addition, Fig. 8 and Tables 5 – 7 show that HIS-FF performed better in terms of convergence and F1-micro scores for most datasets.

In summary, HIS-RW is preferred for use in large-scale training graphs with extremely high time efficiency requirements; otherwise, HIS-FF is a better choice.

## **Appendix G:** Experiments on training hyperparameters and walk length

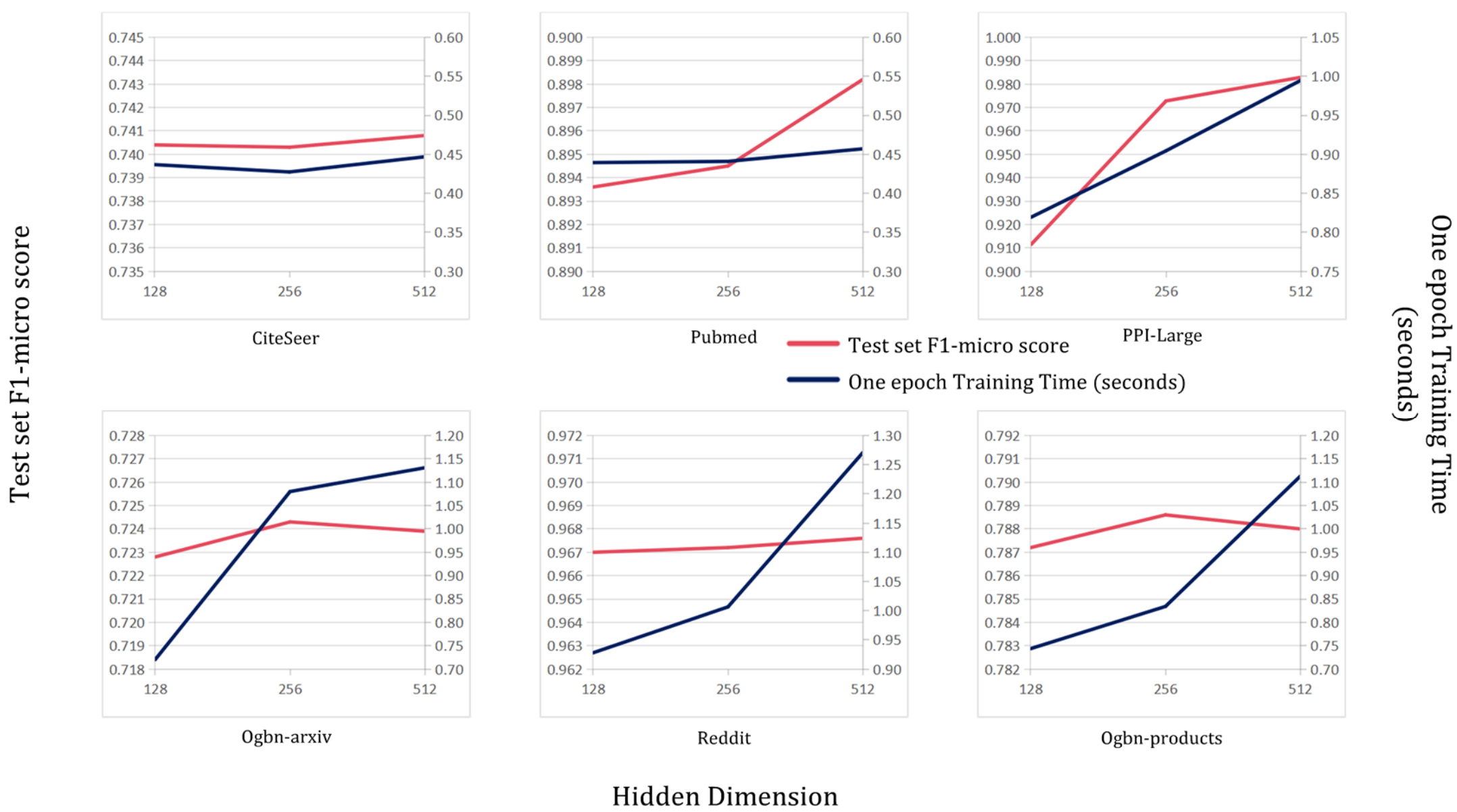

**Figure 11.** Relationship among hidden dimension, test set F1-micro score, and one epoch training time (seconds) of HIS-FF on the same two-layer GraphSAGE model.
The one epoch training time corresponds to the average time of 100 epochs.

In addition to the hyperparameters being analyzed, the other configurations and hyperparameters were the same as those listed in Table 6. Fig. 11 shows the relationship among the hidden dimension, test set F1-micro score, and one epoch training time associated with the HIS-FF sampler. Furthermore, Fig. 12 shows the influences of dropout, learning rate, and walk length (i.e., a sampling parameter of HIS-RW) on test set F1-micro scores. Based on Fig. 12, we found that a high dropout rate is suitable for small-scale training graphs, such as CiteSeer, and the learning rate and the walk length has a relatively weak impact on the training results.

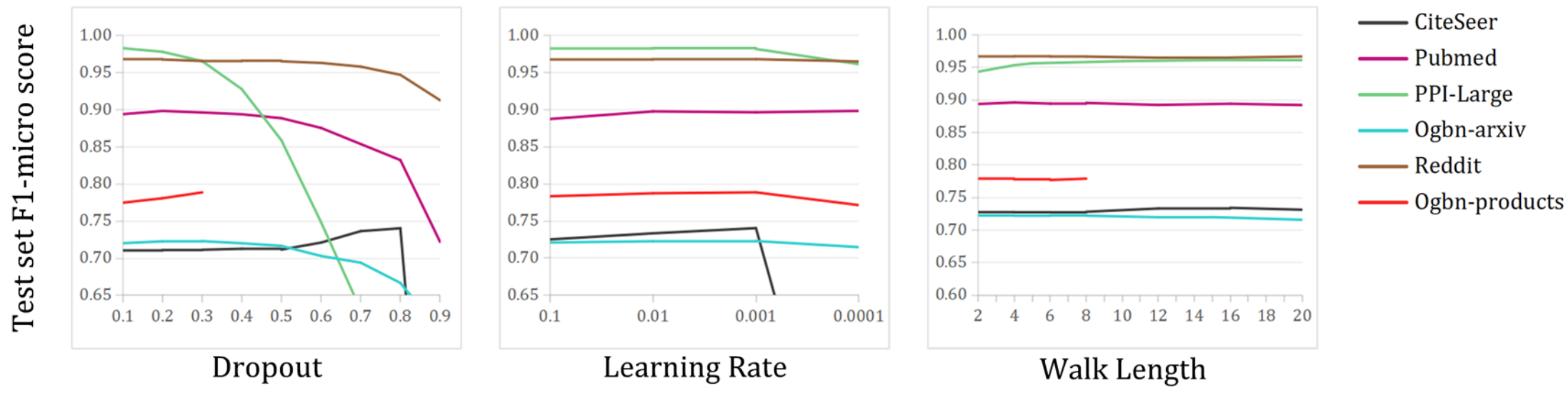

(*a*) (*b*) (*c*)

**Figure 12.** Influences of dropout, learning rate, and walk length on test set F1-micro scores. The experiments in (*a*) and (*b*) adopted the HIS-FF sampler, and the experiments in (*c*) adopted the HIS-RW sampler. All the experiments were executed on the same two-layer GraphSAGE model. For the entire dataset of Ogbn-products with 2,449,026 nodes and 61,859,140 edges, it cannot be accommodated within our limited GPU memory for validation; thus, we performed the training process on the GPU but performed the validation set evaluation on the CPU for the dataset, which leads to higher computational overhead. Owing to time constraints, we did not display the complete curve of the dataset of Ogbn-products in (*a*) and (*c*).

## CRediT authorship contribution statement:

**Qia Hu:** Software; Data curation; Methodology; Formal analysis;
Writing - Original draft.

**Bo Jiao:** Conceptualization; Methodology; Formal analysis; Supervision;
Writing - Review & Editing.

**Corresponding author:** Bo Jiao (jiaoboleetc@outlook.com)